\pdfoutput=1

\documentclass[11pt,dvipsnames]{article}

\usepackage[final]{acl}

\usepackage{times}
\usepackage{latexsym}

\usepackage[T1]{fontenc}

\usepackage[utf8]{inputenc}

\usepackage{microtype}

\usepackage{inconsolata}

\usepackage{graphicx}

\usepackage{hyperref}
\usepackage{url}
\usepackage{booktabs}
\usepackage{orcidlink}
\usepackage{multirow}
\usepackage{xspace}
\usepackage{xcolor}
\usepackage{capt-of}
\usepackage{color}
\usepackage{enumitem}
\usepackage{wrapfig}
\usepackage{comment}
\usepackage{amsmath}
\usepackage{hyperref}

\title{MM-R$^3$: On (In-)Consistency of Vision-Language Models (VLMs)}

\author{Shih-Han Chou$^{1,2}$, Shivam Chandhok$^{1,2}$, James J. Little$^{1}$, Leonid Sigal$^{1,2,3}$\\
$^{1}$Department of Computer Science, University of British Columbia \\ 
$^{2}$Vector Institute for AI \hspace{2mm}
$^{3}$Canada CIFAR AI Chair\\
\texttt{\{shchou75, chshivam, little, lsigal\}@cs.ubc.ca} \\
}

\begin{document}
\twocolumn[{%
\renewcommand\twocolumn[1][]{#1}%
\maketitle
\vspace{-1em}
\begin{center}
    \centering
    \vspace{-1em}
    \captionsetup{type=figure}
    \includegraphics[width=.9\textwidth]{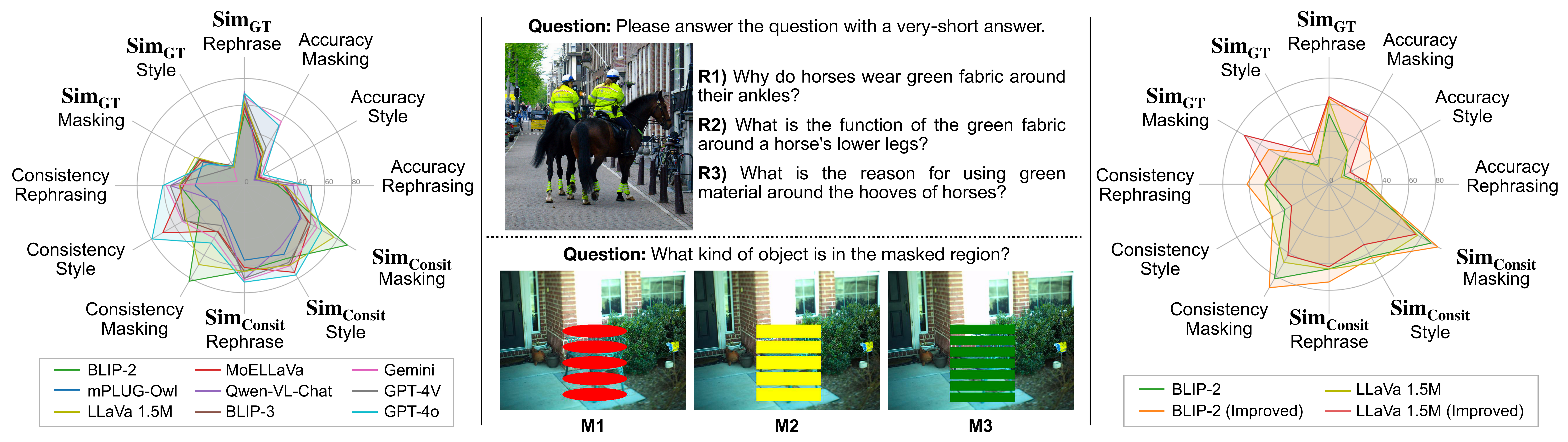}
    \captionof{figure}{(Left) Overall results of VLMs on the MM-R$^3$ Benchmark. (Mid) Consider answering the three semantically identical questions for the top image and a given visual abductive reasoning question for the bottom images from the proposed MM-R$^3$ Benchmark. Humans are accurate and consistent in these tasks while VLMs are much less so. 
    (Right) Results with the proposed adapter.
    }
\label{fig:teaser}
\end{center}%
}]

\begin{abstract}
With the advent of LLMs and 
variants, a flurry of research has emerged, analyzing the performance of such models across an 
array of tasks. While most studies focus on evaluating the capabilities of state-of-the-art (SoTA) Vision Language Models (VLMs) through task accuracy ({\em e.g.}, visual question answering, grounding),
our work explores the related but complementary aspect of {\em consistency} -- the ability of a VLM to produce semantically similar or identical responses to semantically similar queries. We note that consistency is a fundamental prerequisite (necessary but not sufficient condition) for robustness and trust in VLMs.
Armed with this perspective, we propose the \textit{MM-R$^3$ benchmark}, which allows us to analyze performance, in terms of consistency and accuracy, of SoTA VLMs on three tasks: Question Rephrasing, Image Restyling, and Context Reasoning. Our analysis reveals that consistency does not always align with accuracy, indicating that models with higher accuracy are not necessarily more consistent, and vice versa. 
Furthermore, we propose a simple yet effective mitigation strategy in the form of an adapter module trained to minimize inconsistency across prompts. With our proposed strategy, we are able to achieve absolute improvements of $5.7\%$ and $12.5\%$, on average on widely used VLMs such as BLIP-2 and LLaVa 1.5M in terms of consistency over their existing counterparts.
\end{abstract}

\section{Introduction}
\vspace{-.8em}
\label{sec:intro}

Vision Language Models (VLMs)~\citep{liu2023llava, li2023blip2, achiam2023gpt, xue2024xgen}, following and often built on top of purely lingual LLM~\citep{brown2020language, touvron2023llama}, have recently emerged as incredible tools for a broad range of visual understanding tasks, spanning captioning~\citep{lin2014microsoft, sharma2018conceptual, chen2015microsoft}, language grounding~\citep{yu2016modeling, kazemzadeh2014referitgame, liu2019clevr}, visual question answering (VQA)~\citep{VQA}, and many others. 
As the number of such models and their capabilities explode, the research community is progressively focusing on benchmarking their capabilities by developing new benchmarks and testing harnesses. 

Notable examples include MM-Bench~\citep{liu2023mmbench}, SEED-Bench~\citep{li2023seed}, MM-Vet~\citep{yu2023mm}, and others that define numerous tasks that capture a broad range of capabilities of such models ({\em e.g.}, instance counting~\citep{fu2023mme}, spatial relation understanding~\citep{yu2023mm}, abductive~\citep{hessel2022abduction} and deductive~\citep{park2020visualcomet} reasoning). 
These benchmarks continue to shed light on abilities and limitations of VLMs by analyzing their {\em accuracy}. 

However, despite significant progress in the analyses of VLMs, {\em consistency}\footnote{The definition that follows is one of strict consistency that we adopt in this paper. We note that such strict consistency may not be required of all VLMs or problems that they may be designed to solve. However, it is a useful a clean definition that we adopt here.} -- 
{\em ability to produce semantically equivalent response with the  semantically equivalent inputs}, remains broadly overlooked. 
This is a fundamental requirement for VLMs to be reliably deployable for most tasks. 

Anecdotally, LLMs and, by extension, VLMs are sensitive to their prompts which led to the widespread practice of prompt engineering. 
This is problematic as the models' outputs may vary with the phrasing of a query rather than its actual intent, which undermines their reliability. 
Consider the example illustrated in Figure~\ref{fig:teaser} (Mid) top: Most humans would realize that while the three questions (i.e R1, R2, and R3) are superficially different, the semantic meaning is the same. Hence even when the correct answer may not perhaps be apparent ({\em i.e.}, ``to be visible''), the same (consistent) answer should be produced. In contrast, asking models like BLIP-2~\citep{li2023blip2} to answer these questions results in varied responses ``to protect them from splinters'',``to protect the horse's legs'',``to make the hooves more visible'' for the three questions considered.
Similarly, it is obvious to humans that the object being masked in Figure~\ref{fig:teaser} (Mid) bottom is the same irrespective of the type of the mask, and that the object in question is a ``bench''
However, Qwen-VL-Chat's responses vary, indicating ``a bench'',``a yellow object'', ``a green wooden slat sign'' for different masks, highlighting the inconsistency in visual modifications.

It may be tempting to equate {\em accuracy} and {\em consistency}, but the relationship is more intricate. 
While it is true that for objective visual tasks ({\em e.g.}, what color is an object), high accuracy will result in high(er) consistency, current VLMs are far from this high accuracy regime. Further, for more subjective visual tasks ({\em e.g.}, abductive reasoning), high accuracy which tends to be measured as being similar to one of the sets of answers, may not lead to high consistency. In general, one can think of consistency as a necessary but not sufficient property of an AI system and one should seek to maximize \underline{both} {\em consistency} and {\em accuracy}. 

In this work, we first present a comprehensive analysis of SoTA VLMs in terms of their consistency. We do so by developing MM-R$^3$ benchmark atop of the VQA task, where we produce both lingual rephrasings of the original questions (by leveraging GPT-3.5) and visual rephrasings of the image (through stylization) and measure both accuracy and consistency of the produced responses. 
We find that SoTA VLMs while often quite competitive in accuracy 
can differ substantially in their consistency of responses. 
For example, mPLUG-Owl2~\citep{ye2023mplugowl2} is much more susceptible to inconsistency when image inputs are perturbed while MoE-LLaVa~\citep{lin2024moe} is more consistent in the change of the visual domain than the lingual domain. 
In addition, we also define an abductive task of predicting the contents of the masked region, where we find models like BLIP-2 and LLaVa 1.5M~\citep{liu2023llava, liu2023improvedllava} are lower in accuracy but have higher consistency. Overall results for VLMs are illustrated in Figure~\ref{fig:teaser} (Left). We believe these findings both benchmark the capabilities of existing models and outline 
future directions and developments in more consistent VLMs and pre-training objectives ({\em e.g.}, see efforts in LLMs~\citep{aggarwal2023let, chen2023universal, jang2023consistency}). 

Toward the latter goal, we propose a simple adapter module based strategy that effectively improves consistency. 
The adapter is flexible and can be added to any existing VLM. It sits between the VLM embedding layer and the frozen LLM decoder. The goal of the adapter is to modify the LLM’s embeddings such that they are invariant to surface form variations in the language prompt / question or the image. 
We test the proposed adapter in widely used VLMs such as BLIP-2 and LLaVa 1.5M models. 
The experiments show that it is able to improve performance on all three tasks in our proposed benchmark (shown in Figure~\ref{fig:teaser} (Right)).

Our contributions are summarized as follows:
\begin{itemize}[leftmargin=5mm]
\setlength\itemsep{0em}
    \item We propose a new consistency benchmark, MM-R$^3$, that enables evaluation of VLM's ability to produce consistent responses to a range of inputs which are identical semantically, but differ in surface form.
    MM-R$^3$ includes three tasks, covering visual and lingual domains.\vspace{-.25em}
    \item We conduct detailed analyses of SoTA VLMs in accuracy and consistency on this benchmark, taking the first step towards consistency in VLMs, which is a fundamental requirement for VLMs to be reliably deployable for most task.\vspace{-.25em}
    \item We propose a simple but effective adapter-based strategy that can be added to any VLM with lightweight training. Experiments on BLIP-2 and LLaVa 1.5M models show that adding the adapter significantly improves performance on our benchmark regarding consistency metrics.
\end{itemize}

\section{Related Work} 
\label{sec:related}

\noindent{\textbf{Vision Language Models (VLMs).}
The study and development of VLMs~\citep{liu2023improvedllava, li2023blip2, Qwen-VL, achiam2023gpt, lin2024moe, chen2023minigptv2, wang2023cogvlm, Emu2, xue2024xgen} has recently seen a surge in popularity. 
Motivated by the impressive achievement made by recent LLMs~\citep{brown2020language, chowdhery2023palm, achiam2023gpt, touvron2023llama, chung2022scaling}, researchers have ventured into augmenting these models with visual capabilities to tackle multimodal tasks more effectively. A pioneering effort in this realm was the Flamingo model~\citep{alayrac2022flamingo}, which integrated the CLIP image encoder with traditional LLMs. This initiative paved the way for the development of VLMs aimed at enhancing multimodal integration. For example, open-source models like LLaVa 1.5M~\citep{liu2023llava, liu2023improvedllava}, BLIP-2~\citep{li2023blip2}, Qwen-VL~\citep{Qwen-VL}, mPlug-Owl2~\citep{ye2023mplugowl2}, MoE-LLaVa~\citep{lin2024moe}, and BLIP-3~\citep{xue2024xgen}.
Meanwhile, proprietary models like GPT-4V~\citep{achiam2023gpt}, Gemini~\citep{team2023gemini}, and GPT-4o~\citep{gpt4o} have demonstrated state-of-the-art performance, pushing the boundaries of research in this field. However, as VLMs become increasingly powerful, ensuring their reliability across both visual and linguistic domains poses a significant challenge. In this work, we aim to establish a new benchmark for evaluating the {\em consistency} of VLMs.

\vspace{0.3em}
\noindent{\textbf{Vision-Language Benchmarks.}
Traditional Vision-Language (VL) benchmarks have predominantly centered on assessing performance in singular tasks, such as VQA~\citep{VQA, balanced_vqa_v2}, OK-VQA~\citep{okvqa}, MSCOCO~\citep{lin2014microsoft}, and Visual Commonsense Reasoning (VCR)~\citep{zellers2019vcr}. While these benchmarks are valuable, they fall short in fully gauging the comprehensive multimodal perception and reasoning capabilities of VLMs. In response to this gap, a new wave of VL benchmarks has been developed~\citep{fu2023mme, liu2023mmbench, yu2023mm, li2023seed, zhang2024mme}, tailored to the complex demands of VLMs. These benchmarks encompass a range of intricate multimodal tasks that necessitate a seamless integration of vision and language skills. For instance, MME~\citep{fu2023mme} measures perceptual and cognitive skills across a total of $14$ sub-tasks, offering a comprehensive measure of an VLM's abilities. 
MME-RealWorld~\cite{zhang2024mme} is the largest manually annotated benchmark to date, focusing on real-world applications with high-resolution images. 
Despite the advancements these benchmarks represent, their primary focus remains on measuring VLMs' performance in terms of accuracy. 
MAD-Bench~\citep{qian2024easy}, on the other hand, explores VLM robustness by examining how models reconcile discrepancies between textual prompts and images. Our work takes a different stance by prioritizing the consistency of VLMs. In contrast to prior works, we propose a novel consistency benchmark that evaluates not just accuracy, but also the consistency of models across visual and linguistic domains.

\begin{table*}[!t]
    \scriptsize
    \centering
    \caption{\textbf{Statistics of MM-R$^3$ Benchmark.} We list the number of examples in each task as well as the source dataset from which we collect the examples. Test and Train splits are completely disjoint.}\vspace{-1em}
    \begin{tabular}{c|c|c|c}
        \toprule
        & Task &  $\#$Examples & Source \\
        \midrule
        \parbox[t]{2mm}{\multirow{3}{*}{\rotatebox[origin=c]{90}{Train}}} & Question Rephrasing & 16,894 & InfographicsVQA~\citep{mathew2022infographicvqa}, OKVQA~\citep{okvqa} \\
        & Image Restyling & 27,226 & Google Landmarks v2~\citep{weyand2020google},Indoor Scene~\citep{quattoni2009recognizing}\\
         & Context Reasoning & 30,003 & MSCOCO~\citep{lin2014microsoft} \\
        \midrule
        \parbox[t]{2mm}{\multirow{3}{*}{\rotatebox[origin=c]{90}{Test}}} & Question Rephrasing & 3,516 & InfographicsVQA~\citep{mathew2022infographicvqa}, OKVQA~\citep{okvqa} \\
         & Image Restyling & 5,328 & Google Landmarks v2~\citep{weyand2020google},Indoor Scene~\citep{quattoni2009recognizing}\\
         & Context Reasoning & 4,500 & MSCOCO~\citep{lin2014microsoft} \\
        \bottomrule
    \end{tabular}
    \vspace{-2em}
    \label{tab:statistics}
\end{table*}

\vspace{0.3em}
\noindent{\textbf{Consistency in Language.}
Evaluating consistency in LLMs has emerged as a crucial area of inquiry, with semantic consistency being the most widely used focus in consistency analyses. This concept posits that a model should deliver consistent outcomes in semantically equivalent scenarios~\citep{elazar2021erratum}. Notably, it has been observed that pre-trained language models may yield divergent predictions for masked language tasks when singular objects in queries are replaced with their plural counterparts~\citep{ravichander-etal-2020-systematicity} or queries are paraphrased~\citep{elazar2021erratum}, indicating variability responses to semantically similar inputs.
Building on these findings, a recent study~\citep{jang2023consistency} comprehensively investigated LLM consistency. 
Besides consistency evaluation, to improve the robustness of LLMs, \citet{liu2023pre} proposes low-parameter finetuning methods that show a better out-of-distribution performance for generation and classification tasks. Similarly, \citet{newman2021p} proposes a P-Adapter that captures the factual information from the input prompts and improves the prediction consistency.
Inspired by these pioneering efforts, 
our approach seeks to extend this evaluation to multimodal contexts. We employ three distinct tasks, question rephrasing, image restyling, and context reasoning, to analyze the consistency of responses generated by VLMs. 
Furthermore, we build an adapter to mitigate inconsistency that results from semantically equivalent lingual and visual prompts.

\section{MM-R$^3$ Benchmark}
\label{sec:consistency}

\subsection{Overview of MM-R$^3$ Benchmark} 

We introduce a new benchmark designed to assess the semantic consistency of VLMs across both visual and linguistic domains. 
To achieve this, we have crafted three specific tasks: {\em question rephrasing}, {\em image restyling}, and {\em context reasoning}. These tasks are 
designed to probe the models' ability to maintain consistency in their responses. Comprehensive statistics of our benchmark are presented in Table~\ref{tab:statistics}.
The questions and images utilized in this benchmark are derived from existing datasets, which have been adapted to our tasks, ensuring a thorough evaluation of VLM consistency.

\subsection{Tasks and Data Collection}
We describe the tasks in MM-R$^3$ Benchmark and present two examples (see Figure~\ref{fig:data_example_all}) on each task. 

\begin{figure*}[!t]
\caption{(a) The original question is transformed into three distinct versions, each employing a different questioning style. (b) The original images are converted into four stylistic versions. (c) In the Context Reasoning task, objects in the images are masked using various masking methods to test models’ inferential capabilities.}
\vspace{-.5em}
\label{fig:data_example_all}
\includegraphics[width=\linewidth]{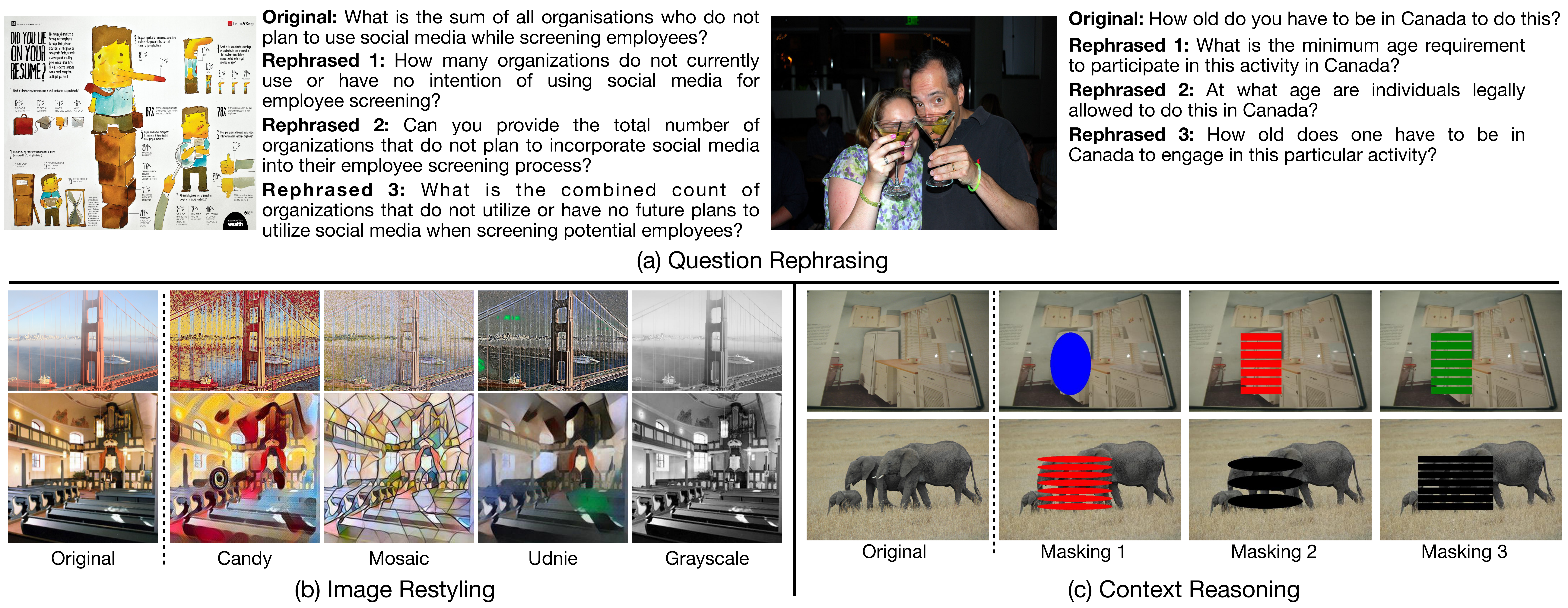}
\vspace{-1em}
\end{figure*}

\vspace{0.3em}
\noindent{\textbf{Question Rephrasing.}}
The goal of the question rephrasing task is to assess the ability of VLMs to provide consistent responses to variously rephrased questions. 
For this task, we employ GPT-3.5 to generate alternate versions of given questions. The questions and images are sourced from the InfographicsVQA~\citep{mathew2022infographicvqa} and OKVQA~\citep{okvqa} datasets. To make the rephrasing meaningful, we take the questions containing more than $10$ words to do the rephrasing. Using the prompt \textit{``I have a question: $\langle$Question$\rangle$ Please give me three different types of rephrased questions to which the answer would be $\langle$Answer$\rangle$.''}, we obtain three distinct rephrasings for each question. These rephrased questions, along with the corresponding images, are then presented to VLMs to evaluate their consistency with respect to linguistic surface form perturbations. 
We randomly select questions from the subset that meet the criteria.
In total, we select $760$ images and derive $3,516$ rephrased questions.

\vspace{0.3em}
\noindent{\textbf{Image Restyling.}}
Humans possess remarkable ability to recognize places and objects in images despite changes in style. We posit that VLMs should demonstrate a similar level of adaptability. To this end, we have devised a task that assesses VLM consistency in the face of varied image styles. 
We choose stylization over more standard image augmentation transformations ({\em e.g.}, TrivialAugment \cite{TrivialAugment} or RandAugment \cite{RandAugment}) to ensure there is no data leakage; simpler data augmentations are often used in training VLMs themselves.
To generate variations of styles, we leverage the style transfer model of~\cite{johnson2016perceptual} along with Instance Normalization~\citep{ulyanov2016instance}. Our dataset comprises both outdoor scenes from the Google Landmarks Dataset v2~\citep{weyand2020google} and indoor scenes from the Indoor Scene Dataset~\citep{quattoni2009recognizing}. The images undergo transformations to exhibit four distinct styles: Candy, Mosaic, Udnie, and Grayscale. The VLMs are then tasked with describing the depicted places in two sentences by using the prompt \textit{``Please describe the place in the image in two sentences.''}, based on these stylistically altered images. This approach allows us to evaluate VLMs consistency with respect to visual domain variations, {\em e.g.}, the ability to describe the scene in a semantically similar manner irrespective of induced image style. 
We randomly collect $600$ outdoor and $732$ indoor scenes from the datasets, resulting in a total of $5,328$ styled images.

\vspace{0.3em}
\noindent{\textbf{Context Reasoning.}}
Spatial context plays an important role in both human and computational perception \cite{Torralba2003}; it allows resoning under uncertainty. 
Testing VLMs on such a task carries a few key benefits. First, the task itself is clearly relevant and important. Second, it is non-deterministic and multiple pleasurable interpretations may rightfully exist. This is precisely a scenario where consistency maybe harder to achieve;  helping to elucidate limitations of current models.  
To test VLMs in this setting, 
we introduce an image masking task where objects within images are randomly masked. We select images from the MSCOCO dataset~\citep{lin2014microsoft}, ensuring that the size of the masked object relative to the entire image falls within a range of $0.1$ to $0.25$. This criterion guarantees that the masked objects are neither too small to be indiscernible nor too large to dominate the image. The masking is applied using three distinct types: lines, shapes, and colors. 
The choice of different types of masks is motivated by wanting to study to
which extent the shape and color of a mask may bias accuracy and consistency of VLM responses.
Specifically, we use $[1,3,5,7]$ lines, rectangle/ellipse shapes, and choose from red/blue/green/yellow/white/black colors for masking. 
Then, we present the masked images to VLMs alongside the query: \textit{``What kind of object is in the masked region?''} 
In total, $1,500$ images are randomly sampled from the subset that meet the constraint, resulting in $4,500$ masked images.

\noindent{\textbf{Data quality validation.}} 
Since validating a large portion of the dataset (with 87,000 samples) manually would be exceedingly costly and time-consuming, we adopt two alternate strategies to evaluate the quality of our dataset: (1) We human validate $300$ randomly sampled question-rephrasing pairs, and $300$ stylized images. We find $93\%$ language rephrasings and $86\%$ image restylings have semantic equivalence for humans; 
(2) we use the InternVL-26B~\cite{chen2024internvl} model (a strong VLM not part of our analysis, with capabilities exceeding GPT4-o in many cases) to automatically validate ALL of the data for the rephrasing task and find it to be $88\%$ semantically equivalent according to InternVL. Note that this is likely a lower bound as InternVL itself is not perfect. 
This validates high quality of our data.

\subsection{Evaluation Method}
The responses expected from VLMs in our tasks are inherently open-ended, reflecting the diverse capabilities of these models. However, the design of our questions is meticulously aligned with the ground truth in the original dataset annotations. This alignment enables us to leverage the annotations effectively during evaluation.
To systematically assess the performance of VLMs, we introduce four distinct evaluation methods, each capturing different aspects of model performance. 

\vspace{0.3em}
\noindent{\textbf{Accuracy ($\mathbf{Acc}$).}}
To evaluate accuracy we assess the responses from VLMs based on an exact match criterion (string matching) with the ground truth annotations. Specifically, if the ground truth annotation is encompassed within the VLM's response, we consider the response correct. 
The accuracy score is then calculated as the average of correct responses across the benchmark dataset. 

\noindent{\textbf{Similarity with GT ($\mathbf{S_{GT}}$).}}
Given the limitations of the exact match criterion, which may inaccurately categorize semantically similar responses as incorrect, we introduce a similarity metric designed to evaluate the nuanced semantic parallels between VLMs' responses and the ground truth annotations. For instance, terms like \textit{person} share semantic similarity with \textit{man} and \textit{woman}, yet would be deemed incorrect under a strict accuracy metric. To address this, our approach utilizes the semantic similarity metric, Sentence-Similarity~\citep{reimers-2019-sentence-bert}, which leverages LLM encodings to assess the similarity between responses and target answers. This metric provides a more subtle assessment of model, rewarding semantic accuracy over mere lexical matching.
The overall performance is the average score across the dataset.

\noindent{\textbf{Consistency Accuracy ($\mathbf{Con}$).}}
This metric is designed to quantify the proportion of responses that manifest a specified degree of semantic consistency. To achieve this, we leverage the same semantic similarity metrics to compute the pairwise similarity scores between responses, utilizing a threshold of $0.7$ to delineate semantic consistency. The threshold is based on the observation of Semantic Textual Similarity benchmark~\citep{cer2017semeval}. A response is deemed consistent if its similarity score with a paired response surpasses this threshold. 
The metric's final value is obtained by averaging the proportion of consistent responses across the entire dataset, providing an aggregate measure of semantic consistency within the VLMs' outputs.

\noindent{\textbf{Consistency Similarity ($\mathbf{S_C}$).}}
Similar to the Consistency Accuracy metric, we measure the consistency similarity by calculating pairwise similarity scores between responses. Instead of setting a threshold, we derive the final metric by computing the average of these similarity scores across the entire dataset. This allows us to more {\em continuously} asses the coherence of responses. 

\vspace{0.3em}
\noindent{\textbf{Overall Performance ($O_{all}$).}}
Motivated by works in generalized few-shot recognition, we propose to use the Harmonic mean ($H_{mean}$) of correctness and consistency as a
combined overall metric for model performance. 
We first calculate the average of $\mathbf{Acc}$ and $\mathbf{S_{GT}}$, two metrics that evaluate correctness against the ground truth. Next, we compute the average of $\mathbf{Con}$ and $\mathbf{S_C}$, two metrics assessing the consistency of generated responses. Finally, we combine these two averages into one single score using the harmonic mean, as we believe this approach can reduce bias when averaging values with large disparities. We use harmonic mean since ideally we want a model to be both correct and consistent and it helps balance these aspects:
\vspace{-.5em}
\begin{equation}
\small
H_{mean}(mean(\mathbf{Acc},\mathbf{S_{GT}}),mean(\mathbf{Con},\mathbf{S_C})).
\end{equation}

\section{Evaluation Results}
\label{sec:results}

\subsection{Models}
All experiments are conducted on NVIDIA A40 GPUs. We evaluate a total of six widely used, open-sourced VLMs, including BLIP-2~\citep{li2023blip2}, mPLUG-Owl2~\citep{ye2023mplugowl2}, LLaVa 1.5M~\citep{liu2023improvedllava, liu2023llava}, MoE-LLaVa~\citep{lin2024moe}, Qwen-VL-Chat~\citep{Qwen-VL}, and BLIP-3~\citep{xue2024xgen} on our consistency benchmark. Additionally, the proprietary models, Gemini~\citep{team2023gemini}, GPT-4V~\citep{achiam2023gpt}, and GPT-4o~\citep{gpt4o}, are included to enrich our comparative analysis. Details of these models are provided in the Appendix~\ref{appx:VLM_details}.

\begin{table*}[!t]
\scriptsize
\begin{center}
\caption{\textbf{Overall results on the MM-R$^3$ Benchmark.} The numbers in red indicate the difference between \textit{Sampling} and \textit{All} results. The best-performing model in each metric is in bold and the second-best model is underlined.}
\begin{tabular}{c|c|l|c|c|c|c|c|c|c|c|c|c}
    \toprule
    & & & \multicolumn{5}{c|}{\textbf{Sampling}} & \multicolumn{5}{c}{\textbf{All}}\\
    \midrule
    & & \textbf{Models} & $\mathbf{Acc}$ & $\mathbf{S_{GT}}$ & $\mathbf{Con}$ & $\mathbf{S_{C}}$ & $O_{all}$ & $\mathbf{Acc}$ & $\mathbf{S_{GT}}$ & $\mathbf{Con}$ & $\mathbf{S_{C}}$ & $O_{all}$ \\
    \midrule
    \parbox[t]{2mm}{\multirow{10}{*}{\rotatebox[origin=c]{90}{\textbf{Question Rephrasing}}}} & \parbox[t]{2mm}{\multirow{6}{*}{\rotatebox[origin=c]{90}{Open-sourced}}} & BLIP-2 & 30.29 & 56.25 & 100.00 & 100.00 & 60.40 & 25.13 \color{Red}{\tiny {(-5.16)}} & 52.91 \color{Red}{\tiny {(-3.34)}} & 48.15 \color{Red}{\tiny {(-51.85)}} & 63.90 \color{Red}{\tiny {(-36.10)}} & 46.00\\
    & & mPLUG-Owl2 & 40.10 & 70.12 & 58.58 & 73.26 & 60.03 & 30.44 \color{Red}{\tiny {(-9.67)}}& 61.10 \color{Red}{\tiny {(-9.03)}} & 36.71 \color{Red}{\tiny {(-21.86)}}& 55.63 \color{Red}{\tiny {(-17.63)}} & 45.97 \\
    & & LLaVa 1.5M & 40.02 & 70.91 & 100.00 & 100.00 & 71.35 & 31.01 \color{Red}{\tiny {(-9.01)}} & 62.85 \color{Red}{\tiny {(-8.06)}} & 48.47 \color{Red}{\tiny {(-51.53)}} & 63.99 \color{Red}{\tiny {(-36.01)}} & 51.16\\
    & & MoE-LLaVa & 34.47 & 65.94 & 81.83 & 87.48 & 63.03 & 28.85 \color{Red}{\tiny {(-5.62)}} & 57.97 \color{Red}{\tiny {(-7.97)}} & 45.32 \color{Red}{\tiny {(-36.51)}} & 61.16 \color{Red}{\tiny {(-26.32)}} & 47.82\\
    & & Qwen-VL-Chat & 52.05 & 82.08 & 93.25 & 95.20 & 78.36 & 36.31 \color{Red}{\tiny {(-15.73)}} & \textbf{70.21} \color{Red}{\tiny {(-11.87)}} & 55.34 \color{Red}{\tiny {(-37.91)}} & 69.70 \color{Red}{\tiny {(-25.51)}} & 57.52\\
    & & BLIP-3 & 32.70 & 55.44 & 80.00 & 80.00 & 56.83 & 30.94 \color{Red}{\tiny {(-1.76)}} & 60.98 \color{ForestGreen}{\tiny {(5.54)}} & 47.70 \color{Red}{\tiny {(-32.3)}} & 63.99 \color{Red}{\tiny {(-16.01)}} & 50.42\\
    \cmidrule{2-13}
    & \parbox[t]{2mm}{\multirow{3}{*}{\rotatebox[origin=c]{90}{Closed-}}} & Gemini & 47.50 & 71.98 & 86.75 & 90.48 & 71.37 & 39.35 \color{Red}{\tiny {(-8.15)}} & 66.22 \color{Red}{\tiny {(-5.76)}} & \underline{58.26} \color{Red}{\tiny {(-28.49)}} & \underline{70.66} \color{Red}{\tiny {(-19.82)}} & 58.04\\
    & & GPT-4V & 53.26 & 64.75 & 76.09 & 81.58 & 67.49 & \textbf{50.22} \color{Red}{\tiny {(-3.04)}} & 67.27 \color{ForestGreen}{\tiny {(2.52)}} & 55.26 \color{Red}{\tiny {(-20.83)}} & 69.18 \color{Red}{\tiny {(-12.41)}} & \underline{60.43}\\
    & & GPT-4o & 57.06 & 76.53 & 85.42 & 89.06 & 75.66 & \underline{46.99} \color{Red}{\tiny {(-10.07)}} & \underline{69.04} \color{Red}{\tiny {(-7.49)}} & \textbf{60.87} \color{Red}{\tiny {(-24.55)}} & \textbf{72.01} \color{Red}{\tiny {(-17.05)}} & \textbf{61.94}\\
    \midrule
    \midrule
    \parbox[t]{2mm}{\multirow{10}{*}{\rotatebox[origin=c]{90}{\textbf{Image Restyling}}}} & \parbox[t]{2mm}{\multirow{6}{*}{\rotatebox[origin=c]{90}{Open-Sourced}}} & BLIP-2 & 16.82 & 16.40 & 100.00 & 100.00 & 28.49 & 13.01 \color{Red}{\tiny {(-3.81)}} & \underline{17.02} \color{ForestGreen}{\tiny {(0.62)}} & 38.36 \color{Red}{\tiny {(-61.64)}} & 62.83 \color{Red}{\tiny {(-37.17)}} & 23.15\\
    & & mPLUG-Owl2 & 15.71 & 14.12 & 53.47 & 69.73 & 24.01 & 8.95 \color{Red}{\tiny {(-6.76)}} & 15.25 \color{ForestGreen}{\tiny {(1.13)}} & 26.29 \color{Red}{\tiny {(-27.18)}} & 59.21 \color{Red}{\tiny {(-10.52)}} & 18.86\\
    & & LLaVa 1.5M & 15.24 & 15.07 & 100.00 & 100.00 & 26.32 & 10.47 \color{Red}{\tiny {(-4.77)}} & 15.49 \color{ForestGreen}{\tiny {(0.42)}} & 50.08 \color{Red}{\tiny {(-49.92)}} & 68.46 \color{Red}{\tiny {(-31.54)}} & 21.30\\
    & & MoE-LLaVa &15.99 & 15.48 & 95.75 & 86.36 & 26.83 & 12.93 \color{Red}{\tiny {(-3.06)}} & 16.60 \color{ForestGreen}{\tiny {(1.12)}} & \underline{70.27} \color{Red}{\tiny {(-25.48)}} & \underline{74.79} \color{Red}{\tiny {(-11.57)}} & \underline{24.54}\\
    & & Qwen-VL-Chat &13.81 & 15.42 & 77.28 & 82.53 & 24.71 & 8.28 \color{Red}{\tiny {(-5.54)}} & 15.73 \color{ForestGreen}{\tiny {(0.31)}} & 23.10 \color{Red}{\tiny {(-54.18)}} & 53.29 \color{Red}{\tiny {(-29.24)}} & 18.26\\
    & & BLIP-3 & 17.12 & 15.08 & 100.00 & 100.00 & 27.73 & 11.92 \color{Red}{\tiny {(-5.20)}} & \textbf{17.31} \color{ForestGreen}{\tiny {(2.23)}} & 51.39 \color{Red}{\tiny {(-48.61)}} & 68.20 \color{Red}{\tiny {(-31.80)}} & 23.49\\
    \cmidrule{2-13}
    & \parbox[t]{2mm}{\multirow{3}{*}{\rotatebox[origin=c]{90}{Closed-}}} & Gemini & 14.47 & 16.55  & 75.85  & 75.88 & 25.75 & \underline{13.68} \color{Red}{\tiny {(-0.79)}} & 16.15 \color{Red}{\tiny {(-0.4)}} & 53.37 \color{Red}{\tiny {(-22.48)}} & 68.95 \color{Red}{\tiny {(-6.93)}} & 23.98\\
    & & GPT-4V & 25.94 & 15.34 & 97.71 & 85.07 & 33.67 & 9.72 \color{Red}{\tiny {(-16.22)}} & 15.90 \color{ForestGreen}{\tiny {(0.56)}} & 52.55 \color{Red}{\tiny {(-45.16)}} & 66.25 \color{Red}{\tiny {(-18.81)}} & 21.07\\
    & & GPT-4o & 16.03 & 17.65 & 96.65 & 84.20 & 28.39 & \textbf{16.31} \color{ForestGreen}{\tiny {(0.28)}} & 16.83 \color{Red}{\tiny {(-0.82)}} & \textbf{79.60} \color{Red}{\tiny {(-17.05)}} & \textbf{77.07} \color{Red}{\tiny {(-7.13)}} & \textbf{27.35}\\
    \midrule
    \midrule
    \parbox[t]{2mm}{\multirow{10}{*}{\rotatebox[origin=c]{90}{\textbf{Context Reasoning}}}} & \parbox[t]{2mm}{\multirow{6}{*}{\rotatebox[origin=c]{90}{Open-Sourced}}} & BLIP-2 & 28.20 & 39.10 & 100.00 & 100.00 & 50.36 & 27.91 \color{Red}{ \tiny {(-0.29)}} & 38.97 \color{Red}{ \tiny {(-0.13)}} & \textbf{82.44} \color{Red}{ \tiny {(-17.56)}} & \textbf{88.80} \color{Red}{ \tiny {(-11.20)}} & 48.10\\
    & & mPLUG-Owl2 & 24.43 & 35.21 & 33.36 & 53.25 & 35.32 & 24.47 \color{ForestGreen}{ \tiny {(0.03)}} & 35.34 \color{ForestGreen}{ \tiny {(0.12)}} & 27.64 \color{Red}{ \tiny {(-5.71)}} & 47.97 \color{Red}{ \tiny {(-5.29)}} & 33.39\\
    & & LLaVa 1.5M & 28.34 & 42.54 & 100.00 & 100.00 & 52.34 & 28.67 \color{ForestGreen}{ \tiny {(0.32)}} & \underline{42.52} \color{Red}{ \tiny {(-0.03)}} & \underline{68.04} \color{Red}{ \tiny {(-31.96)}} & \underline{77.02} \color{Red}{ \tiny {(-22.98)}} & 47.75\\
    & & MoE-LLaVa & 26.13 & 38.91 & 75.49 & 81.08 & 45.95 & 25.16 \color{Red}{ \tiny {(-0.98)}} & 38.47 \color{Red}{ \tiny {(-0.45)}} & 39.40 \color{Red}{ \tiny {(-36.09)}} & 56.86 \color{Red}{ \tiny {(-24.22)}} & 38.31\\
    & & Qwen-VL-Chat& 20.03 & 32.30 & 75.72 & 83.19 & 39.37 & 20.11 \color{ForestGreen}{ \tiny {(0.08)}} & 32.20 \color{Red}{ \tiny {(-0.10)}} & 30.69 \color{Red}{ \tiny {(-45.03)}} & 48.95 \color{Red}{ \tiny {(-34.24)}} & 31.57\\
    & & BLIP-3 & 28.00 & 36.50 & 100.00 & 100.00 & 48.77 & 27.96 \color{Red}{ \tiny {(-0.04)}} & 36.48 \color{Red}{ \tiny {(-0.02)}} & 40.02 \color{Red}{ \tiny {(-50.98)}} & 54.88 \color{Red}{ \tiny {(-45.12)}} & 38.38\\
    \cmidrule{2-13}
    & \parbox[t]{2mm}{\multirow{3}{*}{\rotatebox[origin=c]{90}{Closed-}}}& Gemini & 55.60 & 57.32 & 68.08 & 78.99 & 63.88 & \textbf{55.31} \color{Red}{ \tiny {(-0.29)}} & \textbf{56.35} \color{Red}{ \tiny {(-0.97)}} & 45.22 \color{Red}{ \tiny {(-22.86)}} & 62.66 \color{Red}{ \tiny {(-16.33)}} & \textbf{54.87} \\
    & & GPT-4V & 33.72 & 20.97 & 37.21 & 58.19 & 34.76 & 32.53 \color{Red}{ \tiny {(-1.19)}}  & 21.25 \color{ForestGreen}{ \tiny {(0.28)}} & 34.37 \color{Red}{ \tiny {(-2.84)}} & 57.05 \color{Red}{ \tiny {(-1.14)}} & 33.86\\
    & & GPT-4o & 52.10 & 31.82 & 59.71 & 70.50 & 51.03 & \underline{51.73} \color{Red}{ \tiny {(-0.37)}} & 31.67 \color{Red}{ \tiny {(-0.15)}} & 49.49 \color{Red}{ \tiny {(-10.22)}} & 66.82 \color{Red}{ \tiny {(-3.68)}} & \underline{48.57}\\
    \bottomrule
\end{tabular}
\label{tab:all}
\end{center}
\end{table*}

\subsection{Main Results}

The main results for different tasks are shown in Tables~\ref{tab:all}. For each task, we present results under Sampling and All Data. \textit{Sampling} denotes querying the VLM multiple times with the identical input image and question to observe natural variation in responses without changing the conditions. This allows us to measure how much difference comes from the stochasticity of the model versus the rephrasing of the condition. For each example, we query the VLM four times and average across the whole dataset. \textit{All} represents the mean performance across the dataset. Additional analysis in terms of consistency for correct vs. incorrect answers is given in Appx~\ref{appx:failure_cases}.

Our key findings present intriguing insights. Notably, we observe a divergence between accuracy and consistency across the three tasks. While accuracy performance remains relatively similar and competitive among the different models, there is a marked disparity in their consistency performances. Moreover, we note significant variability in consistency across models, in contrast to the more stable accuracy performance. Furthermore, the variations in consistency are significantly more substantial in the image restyling and context reasoning tasks, suggesting that changes within the visual domain have a greater influence on consistency than linguistic modifications.

\subsection{Detailed Analysis}

\noindent \textbf{Question Rephrasing.} We provide quantitative results in Table~\ref{tab:all}. In the question rephrasing task, GPT-4V achieves the highest accuracy among the nine evaluated models, with Qwen-VL-Chat outperforming the rest of the open-source models. BLIP-2 and LLaVa 1.5M, set with a temperature of $0$, achieve perfect consistency in sampling at $100\%$. However, their performance significantly drops when faced with rephrased questions, indicating a lower consistency in adapting to question variations compared to other models. It implies that BLIP-2 and LLaVa 1.5M might be sensitive to the input prompts. Another interesting observation is that, although Qwen-VL-Chat ranks lower in accuracy compared to GPT-4V, it is better in consistency metrics.

\vspace{.3em}    
\noindent \textbf{Image Restyling.} The results are shown in Table~\ref{tab:all}. In the image restyling task, GPT-4o stands out across three metrics and MoE-LLaVa leads among open-sourced models. Although BLIP-2 outperforms other models in accuracy, its consistency is exceptionally poor.

\vspace{.3em}      
\noindent \textbf{Context Reasoning.} Table~\ref{tab:all} presents the performance of various VLMs in the image masking task. Among the evaluated VLMs, the BLIP-2 and LLaVa 1.5M models achieve a better performance in consistency metrics. 
On the other hand, the Qwen-VL-Chat and BLIP-3 model show the weakest performance across the VLMs, with a notable drop in consistency metrics, which possibly means a lesser capability for abductive reasoning compared to its counterparts.
The ground truth for this task is the annotations from the MSCOCO dataset (\emph{e.g.}, ``person'', ``toilet'', ``elephant'', ``orange'', \emph{etc.}). Since $\mathbf{Acc}$ represents the exact match between responses and the ground truth, we use $\mathbf{S_{GT}}$ to complement the limitations of $\mathbf{Acc}$. For example, if the ground truth is ``person'' and the response is ``baseball player'', the $\mathbf{Acc}$ would be $0$ since there is no exact match. However, the $\mathbf{S_{GT}}$ would be $54.4$, reflecting the semantic similarity between the two.
There may indeed be ambiguities in the answers; however, given the size of the mask that correlates with object identity and surrounding context, humans are quite consistent on this task in most cases. We conducted a human study on $100$ examples ($300$ images in total) for the context reasoning task. The scores for $\mathbf{Acc}$/$\mathbf{S_{GT}}$/$\mathbf{Con}$/$\mathbf{S_{C}}$ are $66.0$/$82.0$/$95.0$/$97.4$, which aligns with the hypothesis that humans are quite consistent.

\begin{figure*}[!t]
    \centering
    \includegraphics[width=.88\linewidth]{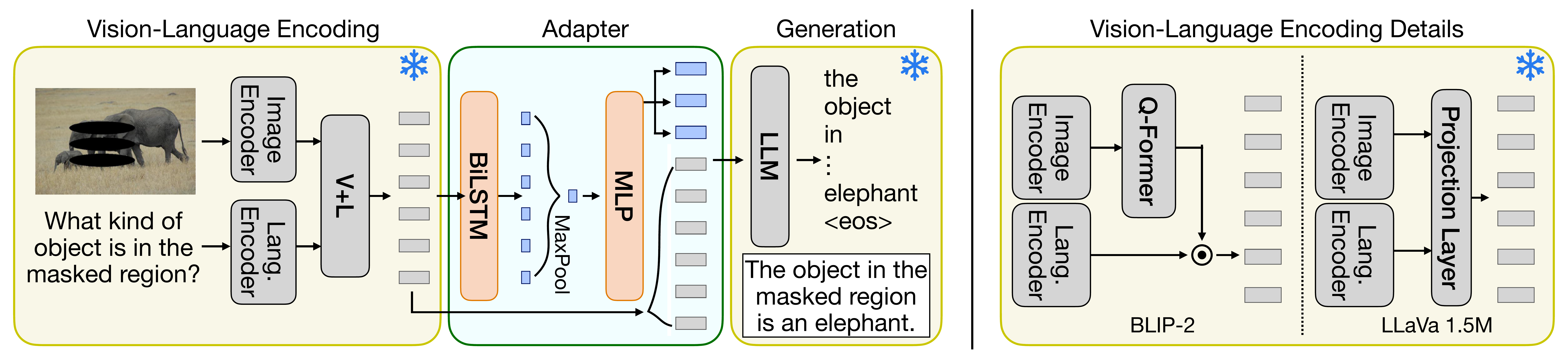}
    \vspace{-.5em}
    \caption{(Left) {\bf Model Overview.} The adapter is added between the Vision-Language Encoding and Generation models. The encoding and decoding parts are frozen during the training of the adapter. (Right) Vision-Language Encoding details of BLIP-2 and LLaVa 1.5M.}
    \vspace{-.5em} 
    \label{fig:model_improvement}
\end{figure*}

\vspace{.3em}  
\noindent
{\bf Analysis Across Tasks.} 
Across three tasks, Qwen-VL-Chat performs better in Question Rephrasing but falls short in Image Restyling and Context Reasoning among the open-sourced models. This disparity may come from its use of the state-of-the-art LLM, Qwen~\citep{qwen}, which likely provides Qwen-VL-Chat with superior initial language representations. On the other hand, BLIP-2 achieves the best performance in Context Reasoning. This is possibly due to its unique image-text learning during pre-training, which could facilitate a better contextual understanding of images. 
We believe the matching between image and language indeed helps the models learn the context in the image. Meanwhile, LLaVa 1.5M and MoE-LLaVa exhibit balanced performances across all tasks, achieving a good trade-off between accuracy and consistency.
Among the closed-source models, GPT-4o outperforms Gemini and GPT-4V in all three tasks. 
However, a notable observation is that current top-performing VLMs still drop significantly on consistency when facing changes in visual and linguistic domains. The gaps suggest that further effort is needed to enhance VLMs' performance in terms of consistency.

\vspace{.3em}     
\noindent
{\bf Qualitative Results.}
We show some qualitative results of the question rephrasing task in Figure~\ref{fig:qualitative_rephrasing}.
The closed-source models outperform other models in terms of both accuracy and consistency. While the accuracy of the evaluated VLMs may not meet high standards, the similarity between the generated responses and the ground truth maintains a satisfactory level (S$_{GT}$), suggesting that the responses are semantically aligned with the ground truth. Regarding consistency, the majority of VLMs demonstrate the capability to generate semantically similar answers to rephrased questions. More results for question rephrasing, image restlying, and context reasoning are in the Appendix~\ref{appx:qualitatives}.

\begin{figure}[!t]
    \centering
    \includegraphics[width=\linewidth]{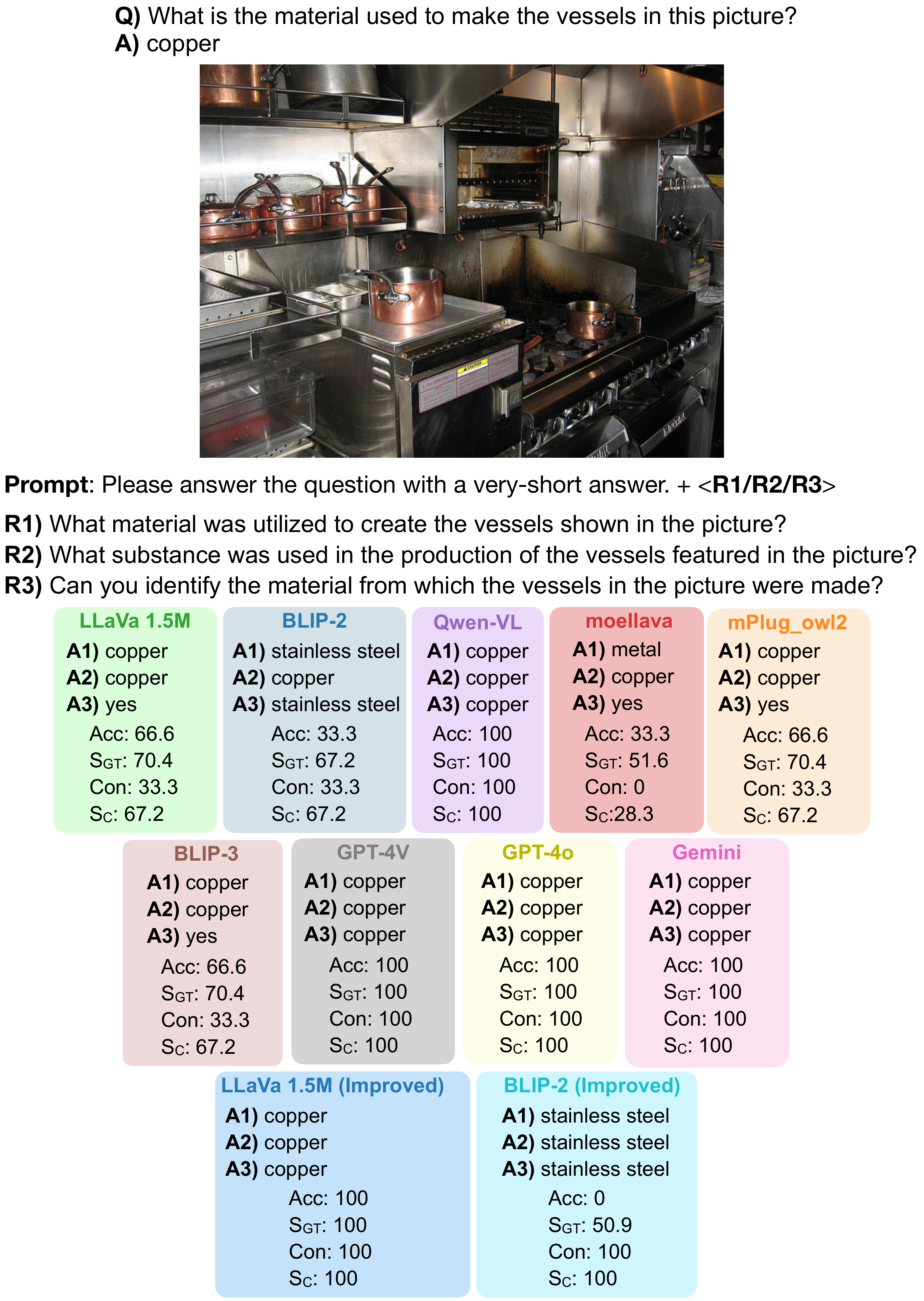}
    \caption{Qualitative Results for the Question Rephrasing Task. \textbf{Q} and \textbf{A} are the original question and answer. \textbf{R1}, \textbf{R2}, and \textbf{R3} denote the rephrased questions.}
    \vspace{-1em}
    \label{fig:qualitative_rephrasing}
\end{figure}

\section{Consistency Improvement}
\label{sec:improvement}
The analyses on the MM-R$^3$ benchmark in Section~\ref{sec:results} show that consistency drops significantly with semantically equivalent prompts.  
To overcome this issue, we propose a simple strategy based on an adapter module that can be plugged into any VLM with lightweight training (Figure~\ref{fig:model_improvement} (green)). The goal of the adapter is to help VLMs overcome the variability of visual and language prompts by making them invariant to surface form variability in data and ensuring a single consistent output.

\begin{table*}[!t]
\scriptsize
\begin{center}
\caption{\textbf{Improvement Results on Three Tasks.} The consistency metrics on three tasks on both BLIP-2 and LLaVa 1.5M models significantly improved after adding the adapter.}
\vspace{-0.5em}
\begin{tabular}{c|l|c|c|c|c|c|c|c|c|c|c|c|c|c|c|c}
    \toprule
    & & \multicolumn{5}{c|}{\textbf{Question Rephrasing}} & \multicolumn{5}{c|}{\textbf{Image Restyling}} & \multicolumn{5}{c}{\textbf{Context Reasoning}}\\
    \midrule
    & \textbf{Models} & $\mathbf{Acc}$ & $\mathbf{S_{GT}}$ & $\mathbf{Con}$ & $\mathbf{S_{C}}$ & $O_{all}$ & $\mathbf{Acc}$ & $\mathbf{S_{GT}}$ & $\mathbf{Con}$ & $\mathbf{S_{C}}$ & $O_{all}$ & $\mathbf{Acc}$ & $\mathbf{S_{GT}}$ & $\mathbf{Con}$ & $\mathbf{S_{C}}$ & $O_{all}$\\
    \midrule
    \parbox[t]{2mm}{\multirow{2}{*}{\rotatebox[origin=c]{90}{Ori.}}} & BLIP-2 & 25.1 & 52.9 & 48.2 & 63.9 & 46.0 & 13.0 & 17.0 & 38.4 & 62.8 & 23.1 & 27.9 & 39.0 & 82.4 & 88.8 & 48.1\\
    & LLaVa & 26.9 & 59.2 & 32.5 & 53.8 & 43.1 & 9.6 & 14.9 & 19.0 & 56.9 & 18.5 & 20.1 & 28.9 & 25.9 & 42.3 & 28.5\\
    \midrule
    \midrule
    \parbox[t]{2mm}{\multirow{2}{*}{\rotatebox[origin=c]{90}{Adapt}}} & BLIP-2 & 24.1 & 62.1 & 58.0 & 71.2 & 51.7 & 27.0 & 27.6 & 49.1 & 66.7 & 37.1 & 53.5 & 51.8 & 88.5 & 94.0 & 66.8\\
    & LLaVa & 28.7 & 64.4 & 42.5 & 61.1 & 49.0 & 20.6 & 25.9 & 32.7 & 52.1 & 30.0 & 55.3 & 71.9 & 62.2 & 74.8 & 66.0\\
    \bottomrule
\end{tabular}
\vspace{-2.5em}
\label{tab:improvement}
\end{center}
\end{table*}

\subsection{Approach}
The design of the adapter is inspired by~\cite{newman2021p}. The adapter model takes the embeddings of the input prompts and outputs a new sequence of continuous embeddings that are used as input to the LLM decoder for generation. Specifically, the adapter takes the outputs of the Vision-Language encoder, passes them through a Bi-LSTM module and a max-pooling procedure to obtain the global embeddings from the input tokens. Then, an MLP is added afterward to project the max-pooling output to the size of the prefix that will be added in front of the original embeddings. In this way, the adapter not only captures the semantic embeddings from the prompts but also retains all of the original embeddings for the LLM decoder.
During fine-tuning, we update only the parameters of the adapter while freezing the vision-language encoder and the language decoder. As shown in Figure~\ref{fig:model_improvement}, the yellow parts are frozen, and the green one represents the training component.

\subsection{Experiments}
\noindent{\textbf{Training Data.}}
We leverage the same data generation pipeline as the MM-R$^3$ benchmark to generate training data samples. Note that these training samples are completely disjoint with the samples in MM-R$^3$ benchmark. In total, there are $16,894$ rephrased question-image pairs for the Question Rephrasing Task, $27,226$ styled images for the Image Restyling Task, and $30,003$ masked images for the Context Reasoning Task for training.

\vspace{.3em}  
\noindent{\textbf{Implementation Details.}}
We select BLIP-2 and LLaVa 1.5M for consistency improvement experiments since they are widely used in recent works, have low consistency compared to other models across lingual changes and allow us to show efficacy of our approach on different types of VLMs families (i.e, use only CLIP vs Q-former based architectures). More details are given in Appx~\ref{appx:consistency_details}.

\vspace{.3em}  
\noindent{\textbf{Results.}}
The performance with and without our adapter, on all three tasks, is illustrated in Table~\ref{tab:improvement}.
In {\em question rephrasing} the accuracy of the model with the adapter is marginally better (improvement of +1.8 for LLaVa 1.5M), however, the consistency shows a very significant margin of improvement (+9.8 and +10 respectively). 
This is consistent with our earlier observation that accuracy and consistency are not necessarily aligned. The original VLMs are already well trained for VQA tasks and so large accuracy boost is not expected. 
However, they are lacking in consistency (see Section~\ref{sec:results}), hence the large improvement on those metrics. 
For {\em image restyling} and {\em context reasoning} we do see a large improvement on both the accuracy and consistency. This is largely because original VLMs are not trained on data of this form. Hence, the introduced adapter can both fine-tune performance on the new data {\em and} improve consistency on them at the same time; 
as shown in the Figure~\ref{fig:qualitative_rephrasing}.

To validate our adapter has not degraded original VLM models, we evaluate on OKVQA dataset before and after the adapter-based fine-tuning. The Original LLaVa 1.5M model shows $58.04$ in Acc and the finetune one is $57.12$. This indicates the proposed adapter significantly improves consistency, with minor, if at all, impact on capabilities.

\section{Conclusion}
\label{sec:conclusion}
In this paper, we explore and analyze {\em consistency} of VLMs and its relationship to {\em accuracy}. 
We do so by introducing
the MM-R$^3$ Benchmark, comprising three tasks -- Question Rephrasing, Image Restyling, and Context Reasoning -- to conduct a comprehensive analysis of SoTA VLMs in terms of both accuracy and consistency. Our analysis reveals that higher accuracy does not necessarily equate to greater consistency in models, and vice versa. In addition, we observe significant variations in the consistency of SoTA models, while their accuracy levels tend to be more stable across models. These findings suggest that future VLM development and objectives might benefit from a dual focus, emphasizing the optimization of consistency and the improvement of overall accuracy. Furthermore, we propose an effective adapter to improve consistency performance. Experiments on BLIP-2, LLaVa
validate improvements.

\vspace{0.05in}
\noindent
{\bf Acknowledgments.}
This work was funded, in part, by the Vector Institute for AI, Canada CIFAR AI Chairs, NSERC Canada Research Chair (CRC), and NSERC Discovery Grants. Resources used in preparing this research were provided, in part, by the Province of Ontario, the Government of Canada through CIFAR, the Digital Research Alliance of Canada\footnote{\url{ https://vectorinstitute.ai/\#partners}}, companies sponsoring the Vector Institute, and Advanced Research Computing at the University of British Columbia. Additional hardware support was provided by John R. Evans Leaders Fund CFI grant and Compute Canada under the Resource Allocation Competition award.

\section{Limitations}
In this work, we focus on semantic input perturbations in both visual and linguistic modalities, as well as semantic output equivalence. Achieving and quantifying this is challenging. Our goal is to develop automatic metrics to simplify the evaluation process. However, the methods for evaluating consistency are not limited to the metrics we propose. For example, some studies use LLMs to assess whether responses are consistent, and future work could further explore this approach with more powerful LLMs.

Additionally, as defined in the introduction, we consider consistency to be a measure of average performance under semantic perturbations, though alternative definitions may also exist. We believe this is an important future direction, where datasets can be expanded and the definition of consistency refined to better encompass a wider range of scenarios.

\bibliography{custom}

\clearpage
\appendix
\section{Appendix}
\label{sec:appendix}

In the appendix, we present additional details and results to complement the main paper.

\subsection{Adversarial robustness vs. consistency.}\label{appx:adver}
The adversarial robustness is related to but different from the consistency we study in this paper.
First, most adversarial robustness approaches~\citep{carlini2017towards, croce2020reliable, ilyas2019adversarial, zhang2019theoretically, bhojanapalli2021understanding} operate in classification settings. Models~\citep{jin2020bert, zhu2023promptbench, moradi2021evaluating} operate on LLMs that have no entitlement of vision and language, and \cite{li2020closer} only studies CLIP (which is a particularly simple contrastive VLM variant). Notably, none of the models deal with VLMs with continuous text outputs. 
Second, they assume the presence of an adversary agent that attempts to find small, local, and often imperceptible, perturbations to inputs (e.g., \cite{bhojanapalli2021understanding} propose pixel level noise perturbations for vision models; \cite{zhu2023promptbench} propose typos and synonyms for LLMs; \cite{moradi2021evaluating} propose character and word level deletions, repetition, etc.), that “produce an incorrect response”~\citep{zhu2023promptbench} or a “decrease in overall classification performance”~\citep{li2020closer}. In other words, robustness is closely tied to accuracy; i.e., robustness only makes sense in the context of a capable model, for samples that the original model is able to classify correctly.
Third, adversarial robustness models, particularly those that attempt to provide theoretic guarantees, quantify worst-case performance under an adversary attack.
In contrast, in studying consistency in VLMs, we focus on a broad class of VLM models that produce open-world textual outputs (including both open- and closed-sourced); this is well beyond CLIP discussed in~\cite{li2020closer} (which is the closest among suggested citations).
Also, we focus on semantic input perturbations (rephrasing and restyling) of both visual and lingual modalities and semantic output equivalence. This is much harder to achieve and quantify. This also goes significantly beyond local word/character perturbations in LLMs or pixel noise perturbations in vision robustness literature.
Importantly, the notion of consistency is entirely devoid of the accuracy or correctness of the original model. Specifically, we study consistency for both all responses and specifically failure cases (see supplementals). A model can be trivially consistent by always responding with the same phrase, irrespective of the input, however, such a model would not generally be considered either accurate or robust under most standard definitions of those two properties. Further, consistency does not assume an adversary, but rather a cooperative agent. In other words, the only perturbations we consider are those likely to be generated by a “typical” user (not one that tries to fool a model). Overall, consistency does not guarantee robustness.
On the other hand, a robust model may also not necessarily guarantee consistency, because typical robustness measures ability for an adversary to flip the decision from correct to incorrect. In more complex tasks (e.g., VQA, captioning), there may be multiple correct answers and also many ways to be incorrect. Consistency measures semantic equivalency even within these classes, which robustness typically does not.
Finally, consistency as we define it, is a measure of average performance under semantic perturbation, not one of worst-case performance.

\subsection{Data Generations and Tasks Pipeline}
We provide a comprehensive pipeline that details both the data collection process and the methods of prompting VLMs for each task, shown in Figure~\ref{fig:Task_pipeline}.

\begin{figure*}[h]
    \centering
    \includegraphics[width=.9\linewidth]{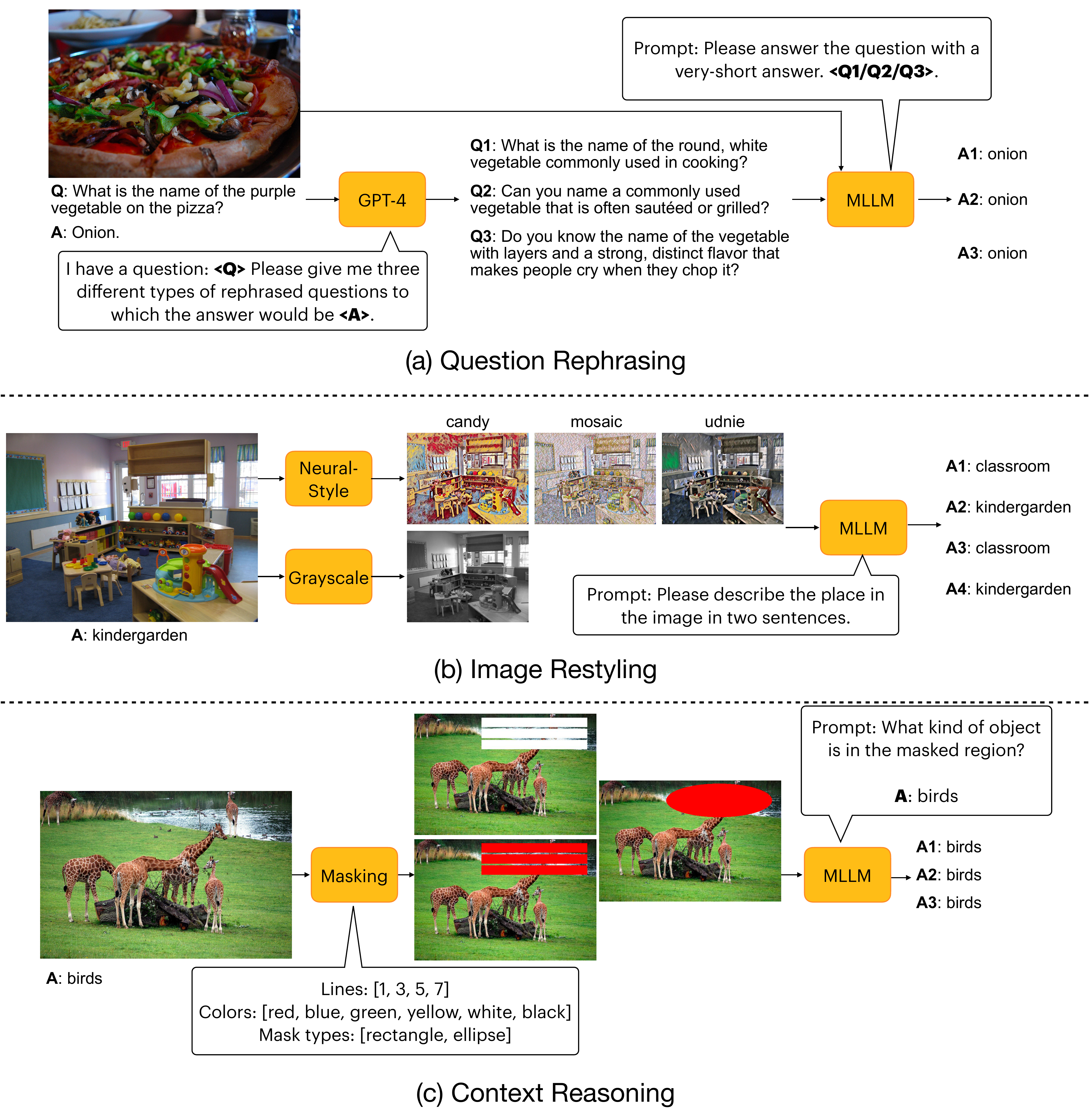}
    \vspace{-1em}
    \caption{The pipeline for the data collection process and prompting VLMs for each task.}
    \label{fig:Task_pipeline}
\end{figure*}

\subsection{Details of VLMs}\label{appx:VLM_details}
\noindent\textbf{BLIP-2}~\citep{li2023blip2}: The BLIP-2 model addresses the modality gap through a novel lightweight Querying Transformer, the Q-former, leveraging a two-stage pre-training approach. Despite its lack of multimodal instruction tuning, BLIP-2 retains the LLM's capacity for following instructions. For our evaluations, we employed the \texttt{BLIP-2 FlanT5XXL} in our evaluations.

\noindent\textbf{mPLUG-Owl2}~\citep{ye2023mplugowl2}: mPLUG-Owl2 utilizes a modularized network design tailored for handling multi-modal inputs. It contains a modality-adaptive module to align different modalities into a shared semantic space for relational learning. The model's architecture, including the visual encoder, visual abstractor, and language decoder, are all updated during training. We use \texttt{mplug-owl2-llama2-7b} for inference and our performance analysis.

\noindent\textbf{LLaVa 1.5M}~\citep{liu2023improvedllava, liu2023llava}: LLaVa is an end-to-end model combining LLaMA/ Vicuna on GPT-generated multimodal instruction-following data. It provides general-purpose visual and language understandings, boasting chat capabilities that parallel the multimodal proficiency of GPT-4. We conducted our experiments using the \texttt{llava-v1.5-7b} version.

\noindent\textbf{MoE-LLaVa}~\citep{lin2024moe}: 
The MoE-LLaVa model incorporates a Mixture of Experts (MoE) architecture with learnable routers, comprising multiple sparse paths that uniquely activate only the top-k experts during deployment while keeping the remaining experts inactive. This design allows MoE-LLaVa to achieve performance comparable to other high-end VLMs with the provided sparse path. We use \texttt{MoE-LLaVA-Phi2-2.7B-4e-384} version for evaluation.

\noindent\textbf{Qwen-VL-Chat}~\citep{Qwen-VL}:
Qwen-VL-Chat builds upon the foundation of Qwen-VL, with training that encompasses not just traditional image descriptions and question-answering tasks, but also extends to grounding and text-reading capabilities through the alignment of image-caption-box tuples. The version tested and analized in our experiments is \texttt{Qwen-VL-Chat}.

\noindent\textbf{BLIP-3}~\citep{xue2024xgen}:
BLIP-3 (xGen-MM) consists of a Vision Transformer, a vision token sampler to downsample the image
embeddings, and a pre-trained Large Language Model. BLIP-3 enables higher-resolution images as inputs by using patch-wise encoding. The patch-wise encoding preserves the resolution of the original images by encoding the split image patches separately. Then a perceiver resampler is used to downsample the visual tokens before sending them to the LLM. This design allows BLIP-3 to reduce the sequence length of vision tokens while keeping the higher-resolution images as inputs. We evaluate the Consistency Benchmark on \texttt{xgen-mm-phi3-mini-instruct-r-v1} version.

\noindent\textbf{Gemini}~\cite{team2023gemini}:
Gemini is developed by Google. It is a multimodal model using a Transformer to process various inputs, such as text, images, audio, and video. Unlike models using separate modules to encode different types of data, Gemini uses a shared Transformer. This enables the model to leverage shared reasoning capabilities. We use \texttt{gemini-1.5-flash} for evaluation.

\noindent\textbf{GPT-4V}~\citep{achiam2023gpt}: GPT-4V extends the capabilities of Generative Pre-trained Transformers by integrating visual understanding, enabling it to process and generate content based on both textual and visual inputs. The development represents a significant advancement in AI, making it a versatile tool for a wide range of applications that require the understanding of both text and imagery. We evaluate the Consistency Benchmark on \texttt{gpt-4-vision-preview} version.

\noindent\textbf{GPT-4o}~\citep{gpt4o}:
GPT-4o (“o” for “omni”) is the most advanced model released by OpenAI. It accepts multimodal inputs, e.g. texts and images. We leverage \texttt{gpt-4o} version to evaluate the Consistency Benchmark.

\subsection{Similarity Metrics Details}
For evaluating Similarity with Ground Truth (S$_{GT}$), Consistency Accuracy (Con), and Consistency Similarity (S$_{C}$), we leverage semantic similarity metrics, Sentence-Similarity~\cite{reimers-2019-sentence-bert}. This metric utilizes large language model encodings to compare the semantic content of predicted and target texts. Specifically, Sentence-Similarity achieves this by transforming input texts into embeddings via a pre-trained Transformer model, capturing their semantic differences. The similarity between these embeddings, and thus the texts, is calculated using cosine similarity, providing a measure of their semantic closeness.

\subsection{Post-processing of Answers}
In the Context Reasoning task, we notice that VLMs often repeat phrases from the question, such as ``the masked region is...'' which raises the consistency scores due to these repetitive terms when calculating answer similarity. However, our primary interest lies in identifying the objects behind the masks. To address this, we conduct post-processing on the responses before metric calculation. This involves filtering out common words like \textit{there}, \textit{is}, \textit{of}, \textit{in}, \textit{the}, etc., and phrases frequently repeated from the question, such as \textit{masked region} and \textit{image}. This adjustment allows for a more fair comparison by focusing on the substance of the responses rather than their structural similarities.

\subsection{Evaluation procedure}
\noindent\textbf{Calculation of Accuracy (Acc)}: To clarify the calculation of accuracy, we do case-insensitive substring matching to validate the response. This works because GT responses tend to be single words or short phrases. Consider the example with three different answers (``columbia'', ``north face'', and ``no'') and the ground truth answer is ``north face'', Acc for three answers is 0/100/0. As a result, the average score for this example will be 33.3 as reported.

\noindent\textbf{Calculation of similarity with GT (S$_{GT}$)}: As the exact match criterion has some limitations, i.e. it may inaccurately categorize semantically similar responses as incorrect, we use a similarity metric in the form of Sentence BERT embeddings.

\noindent\textbf{Calculation of Consistency Accuracy (Con)}:  We compute the pairwise similarity scores between responses using Sentence BERT and utilize a threshold of $0.7$ to delineate semantic consistency. Consider again the example with three different answers (``columbia'', ``north face'', and ``no''), since none of these are semantically similar to one another, the pair-wise Sentence BERT scores are $0.27/0.14/0.24$ — all below $0.7$ threshold and resulting in Con of $0$.

\noindent\textbf{Calculation of Consistency Similarity (S$_{C}$)}: We compute the pairwise similarity scores between responses and average them. Using the same example above, the SC score will be $(0.27+0.14+0.24)/3 = 0.21$.

\subsection{Failure Cases}\label{appx:failure_cases}
In Table~\ref{tab:all-failure}, \textit{Failure Cases} means we only focus on incorrect answers under different changes.
\begin{table*}[h]
\scriptsize
\begin{center}
\caption{Failure results of different models.}
\begin{tabular}{c|l|c|c|c|c|c|c|c|c|c}
    \toprule
    & & \multicolumn{3}{c|}{\textbf{Question Rephrasing Task}} & \multicolumn{3}{c|}{\textbf{Image Restyling Task}} & \multicolumn{3}{c}{\textbf{Context Reasoning Task}} \\
    \midrule
    & \textbf{Models} & $\mathbf{S_{GT}}$ & $\mathbf{Con}$ & $\mathbf{S_{C}}$ & $\mathbf{S_{GT}}$ & $\mathbf{Con}$ & $\mathbf{S_{C}}$ & $\mathbf{S_{GT}}$ & $\mathbf{Con}$ & $\mathbf{S_{C}}$\\
    \midrule
    \parbox[t]{2mm}{\multirow{6}{*}{\rotatebox[origin=c]{90}{Open-Sourced}}} & BLIP-2 & 46.44 & 41.75 & 59.33 & 16.87 & 37.15 & 62.01 & 28.68 & 77.78 & 86.41\\
    & mPLUG-Owl2 & 54.58 & 27.96 & 49.25 & 15.23 & 25.46 & 58.90 & 31.05 & 23.29 & 44.81\\
    & LLaVa 1.5M & 56.26 & 39.83 & 57.99 & 15.43 & 48.72 & 67.86 & 33.28 & 62.34 & 73.14 \\
    & MoE-LLaVa & 52.35 & 38.67 & 56.18 & 16.61 & 69.09 & 74.44 & 33.06 & 33.38 & 53.01\\
    & Qwen-VL-Chat & 62.17 & 44.15 & 62.13 & 15.73 & 22.48 & 52.91 & 28.23 & 26.17 & 45.70\\
    & BLIP3 & 55.56 & 41.23 & 59.49 & 17.29 & 50.64 & 67.85 & 31.22 & 34.50 & 50.61\\
    \midrule
    \midrule
    \parbox[t]{2mm}{\multirow{3}{*}{\rotatebox[origin=c]{90}{Closed-}}} & Gemini & 58.22 & 47.47 & 63.32 & 16.13 & 51.89 & 68.52 & 45.24 & 28.63 & 51.14\\
    & GPT-4V & 52.43 & 34.46 & 54.67 & 15.90 & 51.43 & 65.83 & 17.22 & 29.44 & 54.64\\
    & GPT-4o & 59.85 & 47.01 & 62.76 & 16.78 & 78.34 & 76.67 & 25.01 & 41.72 & 64.27\\
    \bottomrule
\end{tabular}
\vspace{-2em}
\label{tab:all-failure}
\end{center}
\end{table*}

\vspace{-.5em}\subsection{Analysis on Model Size}
Previous studies show that the number of parameters in VLMs affects performance in downstream tasks. For example, BLIP-2, the model achieves greater performance when the number of parameters is larger in VQA, Image Captioning, and Image-Text Retrieval \cite{li2023blip2}. A similar trend is found in the LLaVA 1.5M model \cite{liu2023improvedllava}. As a result, we are curious how consistency is impacted with different sizes of models. We evaluate the BLIP-2 and LLaVA 1.5M models with small and large numbers of parameters. As shown in Table~\ref{tab:model_size}, large models outperform small models on $\mathbf{Acc}$ and $\mathbf{S_{GT}}$ in all three tasks. However, consistency metrics ($\mathbf{Con}$ and $\mathbf{S_{C}}$) do not show a similar trend. In the Question Rephrasing and the Image Restyling task, the large model performs better, while the trend is not the same in the Context Reasoning task. Hence, we observe that unlike accuracy, consistency does not always improve with increase in model size.

\begin{table*}[!t]
\scriptsize
\begin{center}
\caption{{\bf Impact of Model Size.} Different VLM sizes on MM-R$^3$ benchmark. The $^*$ denotes that we ignore generated responses that have an empty output.}
\vspace{-0.5em}
\begin{tabular}{l|c|c|c|c|c|c|c|c|c|c|c|c}
    \toprule
     & \multicolumn{4}{c|}{\textbf{Question Rephrasing}} & \multicolumn{4}{c|}{\textbf{Image Restyling}} & \multicolumn{4}{c}{\textbf{Context Reasoning}}\\
    \midrule
    \textbf{Models} & $\mathbf{Acc}$ & $\mathbf{S_{GT}}$ & $\mathbf{Con}$ & $\mathbf{S_{C}}$ & $\mathbf{Acc}$ & $\mathbf{S_{GT}}$ & $\mathbf{Con}$ & $\mathbf{S_{C}}$ & $\mathbf{Acc}$ & $\mathbf{S_{GT}}$ & $\mathbf{Con}$ & $\mathbf{S_{Con}}$\\
    \midrule
    BLIP-2 (opt2.7B) & 19.0$^*$ & 47.3$^*$ & 39.1$^*$ & 54.3$^*$ & 11.6$^*$ & 55.5$^*$ & 48.5$^*$ & 58.0$^*$ & 27.8 & 39.0 & 76.2 & 86.2 \\ 
    BLIP-2 (flant5xxl) & 25.1 & 52.9 & 48.2 & 63.9 & 13.0 & 17.0 & 38.4 & 62.8 & 27.9 & 39.0 & 82.4 & 88.8\\
    LLaVa 1.5M (7B) & 31.0 & 62.9 & 48.5 & 64.0 & 10.5 & 15.5 & 50.1 & 68.5 & 28.7 & 42.5 & 68.0 & 77.0 \\
    LLaVa 1.5M (13B) & 33.0 & 63.7 & 49.5 & 64.5 & 10.6 & 16.1 & 67.4 & 75.5 & 34.9 & 45.0 & 64.6 & 74.5\\
    \bottomrule
\end{tabular}
\vspace{-2em}
\label{tab:model_size}
\end{center}
\end{table*}

\vspace{-0em}\subsection{Analysis on Different Resolutions in Image Restyling task}\vspace{-.4em}
In the Image Restyling task, we notice that different resolutions of original images might yield different levels of detail when styling the images.
In Image Restyling figure, the Golden Gate has a higher resolution, so it shows more details after restyling. Conversely, the bottom row—the church—has a lower resolution and shows fewer details. To analyze the effect of detail levels for VLMs, we resize the images before performing the style transfer. We resize the images to low ($224\times 224$), mid ($640\times 640$), and high ($1024\times 1024$) resolutions. The results are shown in Figure~\ref{fig:style-resolution} and Table~\ref{tab:style-resolution}. Compared with the original Image Styling task results (Table~\ref{tab:all}), low-resolution results drop significantly, especially in the $\mathbf{Con}$ and $\mathbf{S_{C}}$ metrics, indicating that detail level affects consistency the most. On the other hand, the mid and high-resolution results show consistent improvements, indicating that model consistency increases with an increase in resolution.

\begin{figure*}[!t]
    \centering
    \includegraphics[width=.95\linewidth]{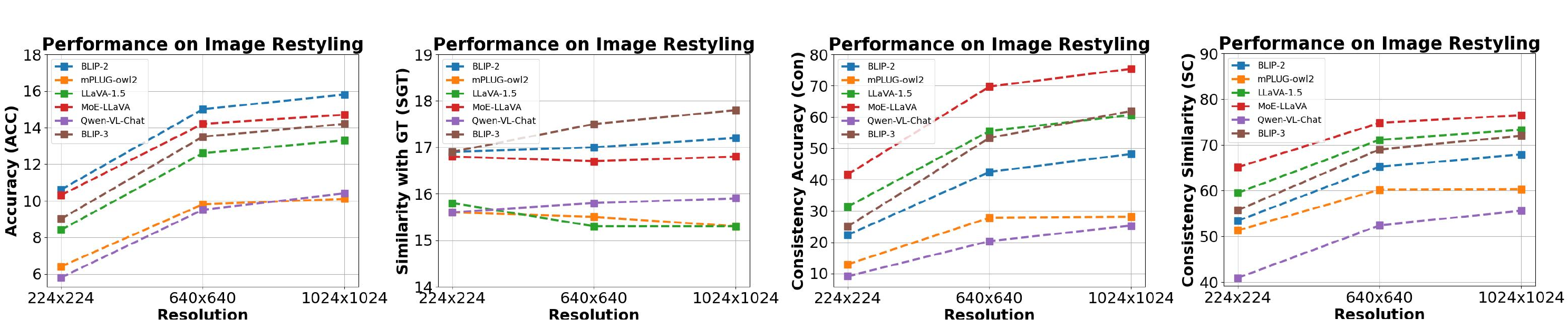}
    \caption{Stylization with different resolutions on the \textbf{Image Restyling Task}.}
    \vspace{-1em}
    \label{fig:style-resolution}
\end{figure*}

\begin{table*}[h]
\scriptsize
\begin{center}
\vspace{-2em}
\caption{Stylization with different resolutions on the \textbf{Image Restyling Task}.}
\begin{tabular}{l|c|c|c|c|c|c|c|c|c|c|c|c}
    \toprule
     & \multicolumn{4}{c|}{\textbf{224$\times$ 224}} & \multicolumn{4}{c|}{\textbf{640$\times$ 640}} & \multicolumn{4}{c}{\textbf{1024$\times$ 1024}}\\
    \midrule
    \textbf{Models} & $\mathbf{Acc}$ & $\mathbf{S_{GT}}$ & $\mathbf{Con}$ & $\mathbf{S_{C}}$ & $\mathbf{Acc}$ & $\mathbf{S_{GT}}$ & $\mathbf{Con}$ & $\mathbf{S_{C}}$ & $\mathbf{Acc}$ & $\mathbf{S_{GT}}$ & $\mathbf{Con}$ & $\mathbf{S_{Con}}$\\
    \midrule
    BLIP-2 & 10.6 & 16.9 & 22.3 & 53.4 & 15.0 & 17.0 & 42.4 & 65.2 & 15.8 & 17.2 & 48.2 & 67.9 \\
    mPLUG-Owl2 & 6.4 & 15.6 & 12.9 & 51.3 & 9.8 & 15.5 & 27.8 & 60.2 & 10.1 & 15.3 & 28.1 & 60.3 \\
    LLaVa 1.5M & 8.4 & 15.8 & 31.3 & 59.5 & 12.6 & 15.3 & 55.5 & 71.1 & 13.3 & 15.3 & 60.6 & 73.3\\
    MoE-LLaVa & 10.3 & 16.8 & 41.6 & 65.1 & 14.2 & 16.7 & 69.7 & 74.8 & 14.7 & 16.8 & 75.4 & 76.5\\
    Qwen-VL-Chat & 5.8 & 15.6 & 9.1 & 40.9 & 9.5 & 15.8 & 20.3 & 52.4 & 10.4 & 15.9 & 25.3 & 55.6 \\
    BLIP-3 & 9.0 & 16.9 & 25.0 & 55.7 & 13.5 & 17.5 & 53.3 & 69.0 & 14.2 & 17.8 & 61.9 & 72.0 \\
    \bottomrule
\end{tabular}
\label{tab:style-resolution}
\end{center}
\end{table*}

\vspace{-.5em}\subsection{Analysis on Different entropy parameters}
The temperature parameter in VLMs controls the level of randomness in the model's output. Lower temperature yields more deterministic outputs, and higher temperature shows more diversity outputs. Hence, we analyze how temperature affects the consistency output in three tasks. We show the results on $3$ different temperatures in Figure~\ref{fig:diff_entrophy} and the actual numbers in Table~\ref{tab:diff_entrophy}. We set the temperatures to $0.2$, $0.7$, and $1$ on all open-sourced models. We notice that mPLUG-Owl2 and MoE-LLaVa model performances drop significantly when the temperature increases while Qwen-VL-Chat show more consistent results. 

\begin{figure*}[!b]
    \centering
    \includegraphics[width=\linewidth]{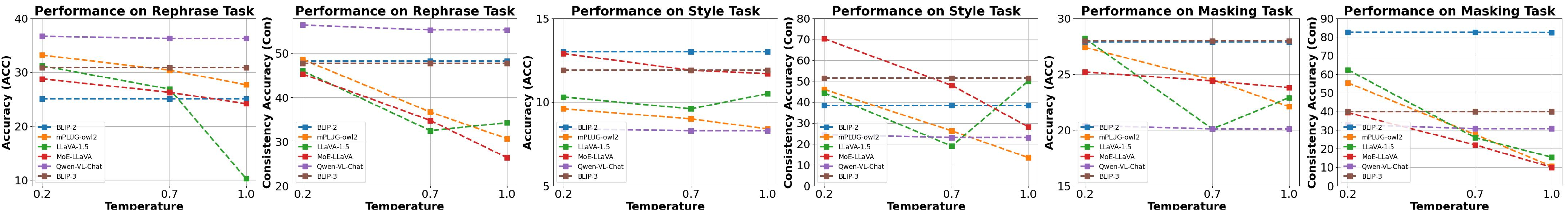}
    \caption{{\bf Impact of Entropy.} Different entropy parameters on all three tasks for models tested.}
    \label{fig:diff_entrophy}
\end{figure*}

\begin{table*}[h]
\scriptsize
\begin{center}
\caption{Different entropy parameters on three tasks.}
\begin{tabular}{l|c|c|c|c|c|c|c|c|c|c|c|c}
    \toprule
    \multicolumn{13}{c}{\textbf{Question Rephrasing Task}} \\
    \midrule
     & \multicolumn{4}{c|}{\textbf{0.2}} & \multicolumn{4}{c|}{\textbf{0.7}} & \multicolumn{4}{c}{\textbf{1}}\\
    \midrule
    \textbf{Models} & $\mathbf{Acc}$ & $\mathbf{S_{GT}}$ & $\mathbf{Con}$ & $\mathbf{S_{C}}$ & $\mathbf{Acc}$ & $\mathbf{S_{GT}}$ & $\mathbf{Con}$ & $\mathbf{S_{C}}$ & $\mathbf{Acc}$ & $\mathbf{S_{GT}}$ & $\mathbf{Con}$ & $\mathbf{S_{Con}}$\\
    \midrule
    BLIP-2 & 25.1 & 52.9 & 48.2 & 63.9 & 25.1 & 52.9 & 48.2 & 63.9 & 25.1 & 52.9 & 48.2 & 63.9\\
    mPLUG-Owl2 & 33.2 & 63.8 & 48.6 & 63.7 & 30.4 & 61.1 & 36.7 & 55.6 & 27.7 & 58.1 & 30.7 & 51.3 \\
    LLaVa 1.5M & 31.2 & 62.6 & 46.0 & 62.4 & 26.9 & 59.2 & 32.5 & 53.8 & 10.3 & 19.3 & 34.3 & 62.3 \\
    MoE-LLaVa & 28.8 & 58.0 & 45.3 & 61.2 & 26.3 & 56.9 & 34.8 & 54.7 & 24.2 & 54.5 & 26.4 & 49.4\\
    Qwen-VL-Chat & 36.7 & 70.6 & 56.4 & 70.4 & 36.3 & 70.2 & 55.3 & 69.7 & 36.3 & 70.2 & 55.3 & 69.7 \\
    BLIP-3 & 30.9 & 61.0 & 47.7 & 64.0 & 30.9 & 61.0 & 47.7 & 64.0 & 30.9 & 61.0 & 47.7 & 64.0 \\
    \midrule
    \midrule
    \multicolumn{13}{c}{\textbf{Image Restyling Task}} \\
    \midrule
    & \multicolumn{4}{c|}{\textbf{0.2}} & \multicolumn{4}{c|}{\textbf{0.7}} & \multicolumn{4}{c}{\textbf{1}}\\
    \midrule
    \textbf{Models} & $\mathbf{Acc}$ & $\mathbf{S_{GT}}$ & $\mathbf{Con}$ & $\mathbf{S_{C}}$ & $\mathbf{Acc}$ & $\mathbf{S_{GT}}$ & $\mathbf{Con}$ & $\mathbf{S_{C}}$ & $\mathbf{Acc}$ & $\mathbf{S_{GT}}$ & $\mathbf{Con}$ & $\mathbf{S_{Con}}$\\
    \midrule
    BLIP-2 & 13.0 & 17.0 & 38.4 & 62.8 & 13.0 & 17.0 & 38.4 & 62.8 & 13.0 & 17.0 & 38.4 & 62.8\\
    mPLUG-Owl2 & 9.6 & 15.4 & 46.1 & 66.6 & 9.0 & 15.2 & 26.3 & 59.2 & 8.4 & 14.9 & 13.5 & 53.8\\
    LLaVa 1.5M & 10.3 & 15.5 & 44.4 & 66.0 & 9.6 & 14.9 & 19.0 & 56.9 & 10.5 & 15.5 & 50.1 & 68.5\\
    MoE-LLaVa & 12.9 & 16.6 & 70.3 & 74.8 & 11.9 & 16.2 & 48.0 & 67.5 & 11.7 & 15.7 & 28.2 & 62.0 \\
    Qwen-VL-Chat & 8.4 & 15.8 & 24.4 & 54.1 & 8.3 & 15.7 & 23.1 & 53.3 & 8.3 & 15.7 & 23.1 & 53.3 \\
    BLIP-3 & 11.9 & 17.3 & 51.4 & 68.2 & 11.9 & 17.3 & 51.4 & 68.2 & 11.9 & 17.3 & 51.4 & 68.2 \\
    \midrule
    \midrule
    \multicolumn{13}{c}{\textbf{Context Reasoning Task}} \\
    \midrule
    & \multicolumn{4}{c|}{\textbf{0.2}} & \multicolumn{4}{c|}{\textbf{0.7}} & \multicolumn{4}{c}{\textbf{1}}\\
    \midrule
    \textbf{Models} & $\mathbf{Acc}$ & $\mathbf{S_{GT}}$ & $\mathbf{Con}$ & $\mathbf{S_{C}}$ & $\mathbf{Acc}$ & $\mathbf{S_{GT}}$ & $\mathbf{Con}$ & $\mathbf{S_{C}}$ & $\mathbf{Acc}$ & $\mathbf{S_{GT}}$ & $\mathbf{Con}$ & $\mathbf{S_{Con}}$\\
    \midrule
    BLIP-2 & 27.9 & 39.0 & 825 & 88.8 & 27.9 & 39.0 & 82.5 & 88.8 & 27.9 & 39.0 & 82.4 & 88.8\\
    mPLUG-Owl2 & 27.4 & 41.0 & 55.4 & 67.0 & 24.5 & 35.3 & 27.6 & 48.0 & 22.1 & 30.1 & 10.5 & 37.2\\
    LLaVa 1.5M & 28.2 & 42.0 & 62.4 & 72.4 & 20.1 & 28.9 & 25.9 & 42.3 & 22.9 & 37.8 & 15.4 & 41.6 \\
    MoE-LLaVa & 25.2 & 38.5 & 39.4 & 56.9 & 24.4 & 34.7 & 22.0 & 45.1 & 23.8 & 31.6 & 10.0 & 38.7\\
    Qwen-VL-Chat & 20.4 & 32.2 & 32.9 & 50.6 & 20.1 & 32.2 & 30.7 & 49.0 & 20.1 & 32.2 & 30.7 & 49.0 \\
    BLIP-3 & 28.0 & 36.5 & 40.0 & 54.9 & 28.0 & 36.5 & 40.0 & 54.9 & 28.0 & 36.5 & 40.0 & 54.9 \\
    \bottomrule
\end{tabular}
\vspace{-2em}
\label{tab:diff_entrophy}
\end{center}
\end{table*}

\subsection{More Analysis}
We provide a deeper analysis of the Image Restyling and Context Reasoning tasks. In the Image Restyling task, we assess performance across $4$ styles: Candy, Mosaic, Udnie, and Grayscale. According to Table~\ref{tab:supp_style}, all VLMs achieve their best results with the \texttt{Grayscale} transformation, which is expected as this style minimally changes the original image. Conversely, the \texttt{Mosaic} style proves challenging for all models, likely due to its significant modification of object appearances, leading to potential confusion for the VLMs.

In the context reasoning task, we assess performance based on masking colors, lines, and shapes. Table~\ref{tab:supp_masking_color} displays the results for various masking colors, with \texttt{Black} outperforming other colors. This could be attributed to the frequent use of black as a bounding box or mask in existing datasets, making VLMs more adept at handling black masks. Table~\ref{tab:supp_masking_lines} examines the effect of different numbers of masking lines, where masks with three lines perform the best, in contrast to those with only one line. This difference may arise from the area covered by the masks or the additional cues multiple lines provide about the underlying objects, aiding VLMs in making predictions. According to Table~\ref{tab:supp_masking_shape}, open-source VLMs show a preference for \texttt{Ellipse} masks over Rectangles, potentially due to the smaller coverage area of ellipses, similar to the influence of the number of lines.

From these observations, it appears that mask color has a minimal impact on performance. Instead, the area covered by the mask plays a more crucial role, suggesting that the inferencing capability of VLMs could be further improved by addressing their sensitivity to the extent of occlusion.

\begin{table*}[!t]
\scriptsize
\begin{center}
\caption{Results across different styles in the \textbf{Image Restyling Task}.}
\begin{tabular}{c|l|c|c|c|c|c|c|c|c}
    \toprule
    &  & \multicolumn{4}{c|}{\textbf{Candy}} & \multicolumn{4}{c}{\textbf{Mosaic}}\\
    \midrule
    & \textbf{Models} & $\mathbf{Acc}$ & $\mathbf{S_{GT}}$ & $\mathbf{Con}$ & $\mathbf{S_{C}}$ & $\mathbf{Acc}$ & $\mathbf{S_{GT}}$ & $\mathbf{Con}$ & $\mathbf{S_{C}}$ \\
    \midrule
    \parbox[t]{2mm}{\multirow{9}{*}{\rotatebox[origin=c]{90}{All}}} & BLIP-2 & 13.59 & 16.73 & 40.99 & 63.80 & 10.51 & 16.24 & 48.35 & 67.91 \\
    & mPLUG-Owl2 & 8.33 & 15.79 & 24.17 & 58.40 & 4.28 & 14.81 & 29.65 & 60.86 \\
    & LLaVa 1.5M & 10.21 & 15.76 & 52.55 & 69.04 & 7.51 & 15.53 & 55.78 & 70.79\\
    & MoE-LLaVa & 13.59 & 17.21 & 69.74 & 74.45 & 9.08 & 17.00 & 74.70 & 76.47 \\
    & Qwen-VL-Chat & 7.51 & 15.88 & 20.27 & 51.69 & 3.45 & 15.00 & 29.35 & 57.96\\ 
    & BLIP-3 & 12.84 & 17.66 & 56.31 & 70.15 & 7.81 & 17.60 & 58.41 & 71.48\\
    & Gemini & 14.11 & 16.58 & 51.88 & 68.63 & 10.21 & 16.50 & 59.01 & 70.75\\
    & GPT-4V & 11.11 & 16.91 & 57.64 & 67.58 & 7.64 & 15.70 & 59.03 & 69.05\\
    & GPT-4o & 15.84 & 17.55 & 81.23 & 77.61 & 12.69 & 17.15 & 86.56 & 79.31\\
    \midrule \midrule
    \parbox[t]{2mm}{\multirow{9}{*}{\rotatebox[origin=c]{90}{Failure Cases}}} & BLIP-2 & - & 16.59 & 39.90 & 63.04 & - & 16.11 & 46.55 & 66.98 \\ 
    & mPLUG-Owl2 & - & 15.76 & 23.50 & 58.09 & - & 14.78 & 28.73 & 60.55 \\
    & LLaVa 1.5M & - & 15.71 & 51.27 & 68.48 & - & 15.48 & 54.05 & 70.11 \\
    & MoE-LLaVa & - & 17.23 & 68.74 & 74.03 & - & 16.97 & 73.51 & 76.13 \\
    & Qwen-VL-Chat & - & 15.91 & 19.59 & 51.28 & - & 15.00 & 28.80 & 57.61\\
    & BLIP-3 & - & 17.69 & 55.52 & 69.81 & - & 17.57 & 58.62 & 71.30\\
    & Gemini & - & 16.58 & 50.16 & 68.20 & - & 16.43 & 57.17 & 70.25\\
    & GPT-4V & - & 16.96 & 56.43 & 67.18 & - & 15.68 & 57.86 & 68.57\\
    & GPT-4o &  & 17.51 & 79.52 & 77.05 & - & 17.02 & 85.63 & 78.93\\
    \bottomrule
    \multicolumn{10}{c}{}\\
    \toprule
    &  & \multicolumn{4}{c|}{\textbf{Udnie}} & \multicolumn{4}{c}{\textbf{Grayscale}}\\
    \midrule
    & \textbf{Models} & $\mathbf{Acc}$ & $\mathbf{S_{GT}}$ & $\mathbf{Con}$ & $\mathbf{S_{C}}$ & $\mathbf{Acc}$ & $\mathbf{S_{GT}}$ & $\mathbf{Con}$ & $\mathbf{S_{C}}$ \\
    \midrule
    \parbox[t]{2mm}{\multirow{9}{*}{\rotatebox[origin=c]{90}{All}}} & BLIP-2 & 10.96 & 16.78 & 36.94 & 62.32 & 16.97 & 18.32 & 40.69 & 63.71 \\
    & mPLUG-Owl2 & 7.96 & 14.99 & 29.43 & 60.53 & 15.24 & 15.41 & 21.92 & 57.31 \\
    & LLaVa 1.5M & 9.01 & 15.36 & 46.17 & 67.11 & 15.17 & 15.32 & 51.20 & 69.87 \\
    & MoE-LLaVa & 12.31 & 16.24 & 68.99 & 74.43 & 16.74 & 15.95 & 70.42 & 74.31 \\
    & Qwen-VL-Chat & 7.43 & 15.45 & 22.37 & 54.13 & 14.71 & 16.57 & 20.95 & 50.74\\ 
    & BLIP-3 & 10.29 & 16.50 & 47.52 & 67.19 & 16.74 & 17.47 & 52.63 & 68.06 \\
    & Gemini & 13.21 & 15.98 & 55.56 & 69.66 & 17.19 & 15.54 & 50.83 & 68.70 \\
    & GPT-4V & 6.25 & 15.59 & 47.22 & 66.38 & 13.89 & 15.38 & 52.08 & 66.11 \\
    & GPT-4o & 16.59 & 16.53 & 77.40 & 76.09 & 20.12 & 16.09 & 79.80 & 77.22\\
    \midrule \midrule
    \parbox[t]{2mm}{\multirow{9}{*}{\rotatebox[origin=c]{90}{Failure Cases}}} & BLIP-2 & - & 16.64 & 35.77 & 61.53 & - & 18.15 & 39.42 & 62.84 \\ 
    & mPLUG-Owl2 & - & 14.97 & 28.65 & 60.22 & - & 15.40 & 20.89 & 56.96 \\
    & LLaVa 1.5M & - & 15.29 & 44.60 & 66.43 & - & 15.24 & 49.84 & 69.31 \\
    & MoE-LLaVa & - & 16.27 & 67.61 & 74.06 & - & 15.99 & 69.47 & 74.06 \\
    & Qwen-VL-Chat & - & 15.43 & 21.66 & 53.79 & - & 16.58 & 20.43 & 50.35 \\
    & BLIP-3 & - & 16.49 & 46.47 & 66.86 & - & 17.43 & 51.95 & 67.75\\
    & Gemini & - & 15.95 & 53.70 & 69.19& - & 15.57 & 49.68 & 68.36\\
    & GPT-4V & - & 15.58 & 45.71 & 66.06 & - & 15.36 & 50.71 & 65.59 \\
    & GPT-4o &- & 16.50 & 75.64 & 75.57 & - & 16.10 & 78.78 & 76.94\\
    \bottomrule
\end{tabular}
\label{tab:supp_style}
\end{center}
\end{table*}

\begin{table*}[!t]
\scriptsize
\begin{center}
\caption{Results across different masking color in the \textbf{Context Reasoning Task}.}
\begin{tabular}{c|l|c|c|c|c|c|c|c|c}
    \toprule
    &  & \multicolumn{4}{c|}{\textbf{Blue}} & \multicolumn{4}{c}{\textbf{Red}}\\
    \midrule
    & \textbf{Models} & $\mathbf{Acc}$ & $\mathbf{S_{GT}}$ & $\mathbf{Con}$ & $\mathbf{S_{C}}$ & $\mathbf{Acc}$ & $\mathbf{S_{GT}}$ & $\mathbf{Con}$ & $\mathbf{S_{C}}$ \\
    \midrule
    \parbox[t]{2mm}{\multirow{9}{*}{\rotatebox[origin=c]{90}{All}}} & BLIP-2 & 26.67 & 38.27 & 82.70 & 89.50 & 27.88 & 39.09 & 79.89 & 87.71 \\
    & mPLUG-Owl2 & 23.51 & 34.62 & 26.07 & 46.06 & 23.06 & 35.26 & 27.88 & 48.38 \\
    & LLaVa 1.5M & 26.31 & 41.51 & 69.18 & 77.44 & 27.88 & 42.22 & 65.82 & 75.70 \\
    & MoE-LLaVa & 20.34 & 35.60 & 37.27 & 56.10 & 23.06 & 37.16 & 35.79 & 54.81 \\
    & Qwen-VL-Chat & 19.85 & 31.78 & 33.50 & 50.14 & 17.69 & 29.97 & 28.95 & 47.80 \\ 
    & BLIP-3 & 23.75 & 35.44 & 39.71 & 54.30 & 28.02 & 37.93 & 40.08 & 54.84 \\
    & Gemini & 54.95 & 56.52 & 44.81 & 62.90 & 51.54 & 54.42 & 43.89 & 61.95 \\
    & GPT-4V & 32.22 & 20.39 & 30.56 & 56.53 & 28.83 & 18.83 & 33.74 & 56.90 \\    
    & GPT-4o & 50.06 & 31.26 & 48.11 & 66.51 & 52.41 & 31.80 & 47.05 & 66.06 \\
    \midrule \midrule
    \parbox[t]{2mm}{\multirow{9}{*}{\rotatebox[origin=c]{90}{Failure Cases}}} & BLIP-2 & - & 27.56 & 78.04 & 87.26 & - & 29.21 & 74.79 & 85.05 \\ 
    & mPLUG-Owl2 & - & 30.55 & 21.21 & 42.66 & - & 30.95 & 24.12 & 45.59 \\
    & LLaVa 1.5M & - & 32.49 & 64.05 & 73.73 & - & 33.10 & 59.93 & 71.72 \\
    & MoE-LLaVa & - & 30.65 & 31.63 & 52.76 & - & 32.01 & 30.04 & 51.16 \\
    & Qwen-VL-Chat & - & 27.21 & 28.76 & 46.66 & - & 26.11 & 24.17 & 44.60 \\
    & BLIP-3 & - & 30.47 & 35.61 & 50.71 & - & 32.45 & 34.56 & 50.19 \\
    & Gemini & - & 45.78 & 28.33 & 51.84 & - & 44.08 & 30.48 & 52.99 \\
    & GPT-4V & - & 17.09 & 26.11 & 54.20 & - & 13.54 & 28.26 & 54.14 \\  
    & GPT-4o & - & 25.15 & 41.40 & 64.35 & - & 24.66 & 39.52 & 63.38 \\
    \bottomrule
    \multicolumn{10}{c}{}\\
    \toprule
    &  & \multicolumn{4}{c|}{\textbf{Green}} & \multicolumn{4}{c}{\textbf{Yellow}}\\
    \midrule
    & \textbf{Models} & $\mathbf{Acc}$ & $\mathbf{S_{GT}}$ & $\mathbf{Con}$ & $\mathbf{S_{C}}$ & $\mathbf{Acc}$ & $\mathbf{S_{GT}}$ & $\mathbf{Con}$ & $\mathbf{S_{C}}$ \\
    \midrule
    \parbox[t]{2mm}{\multirow{9}{*}{\rotatebox[origin=c]{90}{All}}} & BLIP-2 & 27.76 & 39.54 & 85.44 & 90.36 & 28.23 & 38.64 & 83.60 & 89.46 \\
    & mPLUG-Owl2 & 23.13 & 34.06 & 24.90 & 45.90 & 25.00 & 35.72 & 28.23 & 48.94  \\
    & LLaVa 1.5M &  28.30 & 42.14 & 67.48 & 76.50 & 26.88 & 41.57 & 70.03 & 78.57 \\
    & MoE-LLaVa & 22.99 & 36.26 & 39.46 & 56.35 & 19.89 & 34.29 & 39.65 & 57.32 \\
    & Qwen-VL-Chat & 18.78 & 31.92 & 30.20 & 48.70 & 18.68 & 30.69 & 32.12 & 50.19 \\ 
    & BLIP-3 & 23.67 & 36.12 & 41.90 & 55.92 & 24.19 & 34.22 & 38.58 & 54.32 \\
    & Gemini & 57.96 & 57.28 & 45.71 & 62.09 & 52.76 & 55.19 & 45.76 & 63.16 \\
    & GPT-4V & 30.49 & 21.32 & 37.20 & 57.86 & 33.11 & 21.01 & 38.41 & 59.81 \\
    & GPT-4o & 50.88 & 31.49 & 51.70 & 67.16 & 51.21 & 31.64 & 50.54 & 67.32 \\
    \midrule \midrule
    \parbox[t]{2mm}{\multirow{9}{*}{\rotatebox[origin=c]{90}{Failure Cases}}} & BLIP-2 & - & 29.78 & 81.63 & 88.30 & - & 27.43 & 78.70 & 87.08 \\ 
    & mPLUG-Owl2 & - & 32.47 & 61.49 & 72.54 & - & 31.96 & 64.85 & 75.02 \\
    & LLaVa 1.5M & - & 30.20 & 20.66 & 42.89 & - & 30.49 & 23.30 & 45.24 \\
    & MoE-LLaVa & - & 31.10 & 33.84 & 52.82 & - & 28.41 & 33.83 & 53.69 \\
    & Qwen-VL-Chat & - & 28.28 & 26.69 & 45.92 & - & 26.51 & 26.93 & 46.65 \\
    & BLIP-3 & - & 30.16 & 36.70 & 51.96 & - & 28.39 & 32.72 & 50.30 \\
    & Gemini & - & 44.87 & 25.27 & 47.69 & - & 43.47 & 30.10 & 52.20 \\
    & GPT-4V & - & 16.64 & 33.09 & 55.59 & - & 18.47 & 35.82 & 58.54 \\
    & GPT-4o & - & 24.65 & 44.06 & 64.39 & - & 24.43 & 43.00 & 64.77 \\
    \bottomrule
    \multicolumn{10}{c}{}\\
    \toprule
    &  & \multicolumn{4}{c|}{\textbf{White}} & \multicolumn{4}{c}{\textbf{Black}}\\
    \midrule
    & \textbf{Models} & $\mathbf{Acc}$ & $\mathbf{S_{GT}}$ & $\mathbf{Con}$ & $\mathbf{S_{C}}$ & $\mathbf{Acc}$ & $\mathbf{S_{GT}}$ & $\mathbf{Con}$ & $\mathbf{S_{C}}$ \\
    \midrule
    \parbox[t]{2mm}{\multirow{9}{*}{\rotatebox[origin=c]{90}{All}}} & BLIP-2 & 27.40 & 38.14 & 81.92 & 88.02 & 29.59 & 40.14 & 81.11 & 87.64 \\
    & mPLUG-Owl2 & 24.68 & 34.62 & 30.13 & 49.65 & 27.48 & 37.71 & 28.93 & 49.13 \\
    & LLaVa 1.5M & 27.98 & 42.03 & 68.01 & 76.96 & 34.74 & 45.64 & 67.64 & 76.87 \\
    & MoE-LLaVa & 30.42 & 41.84 & 41.61 & 58.18 & 34.87 & 46.00 & 42.93 & 58.54 \\
    & Qwen-VL-Chat & 21.66 & 33.11 & 31.85 & 49.50 & 24.04 & 35.80 & 27.34 & 47.30\\ 
    & BLIP-3 & 31.13 & 37.06 & 41.46 & 56.17 & 37.38 & 38.22 & 38.57 & 53.86\\
    & Gemini & 57.10 & 56.54 & 44.62 & 62.31 & 57.09 & 57.53 & 46.23 & 63.04 \\
    & GPT-4V & 30.20 & 21.81 & 37.58 & 57.89 & 35.95 & 22.86 & 31.37 & 55.80 \\
    & GPT-4o & 53.66 & 32.04 & 50.22 & 67.35 & 52.44 & 31.84 & 49.54 & 66.58\\
    \midrule \midrule
    \parbox[t]{2mm}{\multirow{9}{*}{\rotatebox[origin=c]{90}{Failure Cases}}} & BLIP-2 & - & 28.85 & 77.26 & 85.57 & - & 29.33 & 76.30 & 85.14 \\ 
    & mPLUG-Owl2 & - & 30.26 & 25.60 & 46.48 & - & 33.81 & 25.18 & 46.32 \\
    & LLaVa 1.5M & - & 32.58 & 62.14 & 73.10 & - & 37.15 & 61.33 & 72.64 \\
    & MoE-LLaVa & - & 36.67 & 35.17 & 53.95 & - & 40.16 & 36.20 & 53.80 \\
    & Qwen-VL-Chat & - & 29.05 & 27.30 & 46.17 & - & 32.33 & 23.10 & 44.20 \\
    & BLIP-3 & - & 32.44 & 35.86 & 51.84 & - & 33.58 & 31.55 & 48.78\\
    & Gemini & - & 46.09 & 28.35 & 50.30 & - & 46.44 & 28.63 & 50.75 \\
    & GPT-4V & - & 18.44 & 32.58 & 55.81 & - & 18.44 & 25.78 & 53.19 \\
    & GPT-4o & - & 26.26 & 40.13 & 64.92 & - & 24.94 & 42.16 & 63.82\\
    \bottomrule
\end{tabular}
\label{tab:supp_masking_color}
\end{center}
\end{table*}

\begin{table*}[!t]
\scriptsize
\begin{center}
\caption{Results across different numbers of lines in the \textbf{Context Reasoning Task}.}
\begin{tabular}{c|l|c|c|c|c|c|c|c|c}
    \toprule
    &  & \multicolumn{4}{c|}{\textbf{1}} & \multicolumn{4}{c}{\textbf{3}}\\
    \midrule
    & \textbf{Models} & $\mathbf{Acc}$ & $\mathbf{S_{GT}}$ & $\mathbf{Con}$ & $\mathbf{S_{C}}$ & $\mathbf{Acc}$ & $\mathbf{S_{GT}}$ & $\mathbf{Con}$ & $\mathbf{S_{C}}$ \\
    \midrule
    \parbox[t]{2mm}{\multirow{9}{*}{\rotatebox[origin=c]{90}{All}}} & BLIP-2 & 23.71 & 36.08 & 78.44 & 86.34 & 32.10 & 41.27 & 82.75 & 88.91 \\
    & mPLUG-Owl2 & 19.50 & 31.66 & 25.09 & 45.49 & 29.43 & 38.16 & 30.17 & 48.93 \\
    & LLaVa 1.5M & 23.88 & 38.60 & 61.86 & 72.55 & 33.76 & 46.20 & 69.83 & 78.16\\
    & MoE-LLaVa & 17.27 & 32.56 & 32.99 & 53.31 & 30.54 & 42.35 & 42.99 & 58.78 \\
    & Qwen-VL-Chat & 13.06 & 25.69 & 26.37 & 45.45 & 28.78 & 37.72 & 33.95 & 50.83 \\ 
    & BLIP-3 & 15.81 & 30.66 & 34.19 & 50.24 & 37.82 & 41.47 & 43.82 & 58.42 \\
    & Gemini & 55.98 & 56.17 & 42.73 & 61.62 & 54.24 & 56.60 & 47.51 & 63.40 \\
    & GPT-4V & 35.25 & 20.33 & 32.38 & 55.62 & 33.33 & 21.63 & 34.17 & 58.25 \\   
    & GPT-4o & 46.65 & 29.32 & 48.88 & 66.38 & 56.83 & 33.65 & 49.91 & 66.88 \\
    \midrule \midrule
    \parbox[t]{2mm}{\multirow{9}{*}{\rotatebox[origin=c]{90}{Failure Cases}}} & BLIP-2 & - & 26.47 & 73.36 & 83.62 & - & 30.24 & 77.53 & 86.15 \\ 
    & mPLUG-Owl2 & - & 27.70 & 20.96 & 42.48 & - & 33.19 & 25.05 & 45.25 \\
    & LLaVa 1.5M & - & 29.90 & 55.95 & 68.54 & - & 36.36 & 63.39 & 73.96 \\
    & MoE-LLaVa & - & 27.70 & 27.32 & 49.79 & - & 36.17 & 36.06 & 54.40 \\
    & Qwen-VL-Chat & - & 22.18 & 22.15 & 42.44 & - & 32.94 & 28.18 & 46.68 \\
    & BLIP-3 & - & 26.06 & 29.45 & 46.74 & - & 35.58 & 37.83 & 53.51 \\
    & Gemini & - & 45.94 & 25.78 & 50.20 & - & 44.38 & 29.12 & 50.80 \\
    & GPT-4V & - & 15.74 & 27.67 & 52.91 & - & 18.82 & 30.70 & 56.64 \\   
    & GPT-4o & - & 22.48 & 40.88 & 63.66 & - & 26.96 & 40.42 & 64.06 \\
    \bottomrule
    \multicolumn{10}{c}{}\\
    \toprule
    &  & \multicolumn{4}{c|}{\textbf{5}} & \multicolumn{4}{c}{\textbf{7}}\\
    \midrule
    & \textbf{Models} & $\mathbf{Acc}$ & $\mathbf{S_{GT}}$ & $\mathbf{Con}$ & $\mathbf{S_{C}}$ & $\mathbf{Acc}$ & $\mathbf{S_{GT}}$ & $\mathbf{Con}$ & $\mathbf{S_{C}}$ \\
    \midrule
    \parbox[t]{2mm}{\multirow{9}{*}{\rotatebox[origin=c]{90}{All}}} & BLIP-2 & 29.90 & 40.23 & 84.28 & 89.98 & 26.29 & 38.51 & 84.45 & 90.05 \\
    & mPLUG-Owl2 & 26.02 & 36.44 & 26.92 & 48.63 & 23.32 & 35.34 & 28.56 & 48.92 \\
    & LLaVa 1.5M & 30.08 & 43.75 & 70.82 & 78.99 & 27.34 & 41.81 & 69.96 & 78.56 \\
    & MoE-LLaVa & 28.09 & 40.00 & 39.93 & 57.33 & 25.24 & 39.32 & 42.01 & 58.20 \\
    & Qwen-VL-Chat & 19.87 & 33.27 & 32.25 & 50.75 & 19.30 & 32.57 & 30.48 & 48.98\\ 
    & BLIP-3 & 31.53 & 37.85 & 40.38 & 54.29 & 27.51 & 36.35 & 42.01 & 56.80 \\
    & Gemini & 58.42 & 57.86 & 46.11 & 63.49 & 52.23 & 54.45 & 44.53 & 61.93 \\
    & GPT-4V & 31.19 & 21.95 & 35.32 & 57.87 & 27.52 & 20.22 & 36.82 & 58.01 \\
    & GPT-4o & 54.56 & 33.07 & 49.32 & 67.00 & 49.34 & 30.81 & 49.87 & 67.03 \\
    \midrule \midrule
    \parbox[t]{2mm}{\multirow{9}{*}{\rotatebox[origin=c]{90}{Failure Cases}}} & BLIP-2 & - & 30.10 & 79.95 & 87.98 & - & 28.21 & 80.52 & 88.05 \\ 
    & mPLUG-Owl2 & - & 32.55 & 23.18 & 45.74 & - & 31.04 & 24.13 & 45.87 \\
    & LLaVa 1.5M & - & 34.23 & 65.65 & 75.23 & - & 33.06 & 64.82 & 75.15 \\
    & MoE-LLaVa & - & 34.53 & 34.01 & 53.50 & - & 34.27 & 36.55 & 54.58 \\
    & Qwen-VL-Chat & - & 29.47 & 28.03 & 47.85 & - & 28.91 & 26.67 & 46.08 \\
    & BLIP-3 & - & 32.71 & 34.37 & 49.77 & - & 31.24 & 36.89 & 52.85 \\
    & Gemini & - & 47.07 & 30.64 & 52.25 & - & 43.07 & 29.02 & 51.01 \\
    & GPT-4V & - & 17.66 & 30.16 & 55.79 & - & 16.08 & 32.11 & 55.54 \\
    & GPT-4o & - & 26.55 & 43.16 & 64.69 & - & 24.40 & 42.44 & 64.70 \\
    \bottomrule
\end{tabular}
\label{tab:supp_masking_lines}
\end{center}
\end{table*}

\begin{table*}[!t]
\scriptsize
\begin{center}
\caption{Results across different masking shapes in the \textbf{Context Reasoning Task}.}
\begin{tabular}{c|l|c|c|c|c|c|c|c|c}
    \toprule
    &  & \multicolumn{4}{c|}{\textbf{Rectangle}} & \multicolumn{4}{c}{\textbf{Ellipse}}\\
    \midrule
    & \textbf{Models} & $\mathbf{Acc}$ & $\mathbf{S_{GT}}$ & $\mathbf{Con}$ & $\mathbf{S_{C}}$ & $\mathbf{Acc}$ & $\mathbf{S_{GT}}$ & $\mathbf{Con}$ & $\mathbf{S_{C}}$ \\
    \midrule
    \parbox[t]{2mm}{\multirow{9}{*}{\rotatebox[origin=c]{90}{All}}} & BLIP-2 & 26.51 & 38.32 & 82.66 & 88.98 & 29.22 & 39.58 & 82.24 & 88.63 \\
    & mPLUG-Owl2 & 22.34 & 34.12 & 27.39 & 48.01 & 26.47 & 36.47 & 27.89 & 47.93 \\
    & LLaVa 1.5M & 26.42 & 41.49 & 67.84 & 77.24 & 30.78 & 43.48 & 68.23 & 76.80\\
    & MoE-LLaVa & 20.28 & 35.60 & 39.59 & 56.91 & 29.74 & 41.17 & 39.22 & 56.81 \\
    & Qwen-VL-Chat & 16.33 & 29.24 & 27.89 & 47.01 & 23.66 & 34.99 & 33.32 & 50.77 \\ 
    & BLIP-3 & 24.13 & 35.57 & 40.83 & 55.29 & 31.55 & 37.34 & 39.27 & 54.48 \\
    & Gemini & 54.39 & 56.20 & 45.06 & 62.34 & 55.98 & 56.30 & 45.27 & 62.8 \\
    & GPT-4V & 33.19 & 21.16 & 34.48 & 57.16 & 30.43 & 20.83 & 34.89 & 57.69\\
    & GPT-4o & 47.71 & 30.18 & 50.46 & 67.07 & 55.52 & 33.07 & 48.58 & 66.58\\
    \midrule \midrule
    \parbox[t]{2mm}{\multirow{9}{*}{\rotatebox[origin=c]{90}{Failure Cases}}} & BLIP-2 & - & 27.96 & 78.13 & 86.64 & - & 29.36 & 77.45 & 86.19 \\ 
    & mPLUG-Owl2 & - & 29.94 & 23.06 & 45.00 & - & 32.10 & 23.50 & 44.63 \\
    & LLaVa 1.5M & - & 32.47 & 62.37 & 73.58 & - & 34.05 & 62.31 & 72.71 \\
    & MoE-LLaVa & - & 30.28 & 33.86 & 53.35 & - & 35.71 & 32.93 & 52.68 \\
    & Qwen-VL-Chat & - & 26.04 & 24.26 & 44.36 & - & 30.36 & 28.02 & 47.01\\
    & BLIP-3 & - & 30.24 & 35.64 & 51.27 & - & 32.16 & 33.42 & 49.99\\
    & Gemini & - & 45.49 & 28.64 & 51.24 & - & 44.72 & 28.54 & 50.87\\
    & GPT-4V & - & 16.56 & 29.37 & 54.58 & - & 17.53 & 30.95 & 55.82\\   
    & GPT-4o & - & 23.60 & 42.97 & 64.80 & - & 26.39 & 40.48 & 63.75\\
    \bottomrule
\end{tabular}
\label{tab:supp_masking_shape}
\end{center}
\end{table*}

\subsection{Consistency Improvement Details and More Results}\label{appx:consistency_details}
\noindent{\textbf{Implementation Details.}}
For both BLIP-2 and LLaVa 1.5M experiments, the Bi-LSTM includes $2$ recurrent layers with the hidden size is set to $4096$ and dropout rate $0$. We follow the same setting as in~\cite{newman2021p} to set the prefix size to $3$. The model is optimized by \textit{CrossEntropyLoss}.
The initial linear rate is set to $1e-5$. A linear decay learning rate schedule is also used. We conduct all experiments on a Nvidia A40 GPU with batch size $2$ on both models. The adapter is added on the top of BLIP-2 \texttt{\small blip2-t5 pretrain-flant5xxl} and LLaVa 1.5M \texttt{\small llava-v1.5-7b} models.

\noindent{\textbf{Ablation Studies.}} 
We provide ablation studies in Table~\ref{tab:improvement_ablation}. Adapt (Sep) indicates that the adapter model is trained separately for each task, meaning each task has its own corresponding model. LoRA refers to an alternative approach where, instead of using an adapter to help VLMs handle input variability, we fine-tune the VLMs using LoRA. In the LoRA fine-tuning setting, we not only use the datasets mentioned in the LLaVA 1.5M paper but also incorporate data from the MM$^3$ benchmark.

\begin{table*}[!t]
\scriptsize
\begin{center}
\caption{\textbf{Ablation Studies on Adapter and LoRA.} Adapt (Sep) indicates that the adapter model is trained separately for each task. We also provide the LoRA fine-tuning on LLaVA 1.5M.}
\begin{tabular}{c|l|c|c|c|c|c|c|c|c|c|c|c|c}
    \toprule
    & & \multicolumn{4}{c|}{\textbf{Question Rephrasing}} & \multicolumn{4}{c|}{\textbf{Image Restyling}} & \multicolumn{4}{c}{\textbf{Context Reasoning}}\\
    \midrule
    & \textbf{Models} & $\mathbf{Acc}$ & $\mathbf{S_{GT}}$ & $\mathbf{Con}$ & $\mathbf{S_{C}}$ & $\mathbf{Acc}$ & $\mathbf{S_{GT}}$ & $\mathbf{Con}$ & $\mathbf{S_{C}}$ & $\mathbf{Acc}$ & $\mathbf{S_{GT}}$ & $\mathbf{Con}$ & $\mathbf{S_{C}}$\\
    \midrule
    \multirow{2}{*}{{Ori.}} & BLIP-2 & 25.1 & 52.9 & 48.2 & 63.9 & 13.0 & 17.0 & 38.4 & 62.8 & 27.9 & 39.0 & 82.4 & 88.8 \\
    & LLaVa 1.5M & 26.9 & 59.2 & 32.5 & 53.8 & 9.6 & 14.9 & 19.0 & 56.9 & 20.1 & 28.9 & 25.9 & 42.3\\
    \midrule
    \midrule
    \multirow{2}{*}{Adapt (Sep)} & BLIP-2 & 27.6 & 64.7 & 61.8 & 73.7 & 36.7 & 25.6 & 49.0 & 64.1 & 54.6 & 52.4 & 90.1 & 94.7 \\
    & LLaVa 1.5M & 31.4 & 65.9 & 43.2 & 62.3 & 18.1 & 28.1 & 32.6 & 52.6 & 58.6 & 73.8 & 62.0 & 75.5\\
    \midrule
    \midrule
    \multirow{2}{*}{LoRA} & \parbox[t]{2mm}{\multirow{2}{*}{LLaVa 1.5M}} & \multirow{2}{*}{28.0} & \multirow{2}{*}{62.9} & \multirow{2}{*}{40.9} & \multirow{2}{*}{59.8} & \multirow{2}{*}{27.4} & \multirow{2}{*}{26.1} & \multirow{2}{*}{29.9} & \multirow{2}{*}{51.0} & \multirow{2}{*}{50.5} & \multirow{2}{*}{68.8} & \multirow{2}{*}{56.3} & \multirow{2}{*}{71.6} \\
    & &  &  &  &  &  &   &  &  & & & \\
    \bottomrule
\end{tabular}
\vspace{-2em}
\label{tab:improvement_ablation}
\end{center}
\end{table*}

\subsection{Qualitative Results}\label{appx:qualitatives}

\noindent\textbf{Question Rephrasing}
We show more qualitative results of the question rephrasing
task in Figure~\ref{fig:supp_rephrasing}.

\begin{figure*}[!t]
    \centering
    \includegraphics[width=1\linewidth]{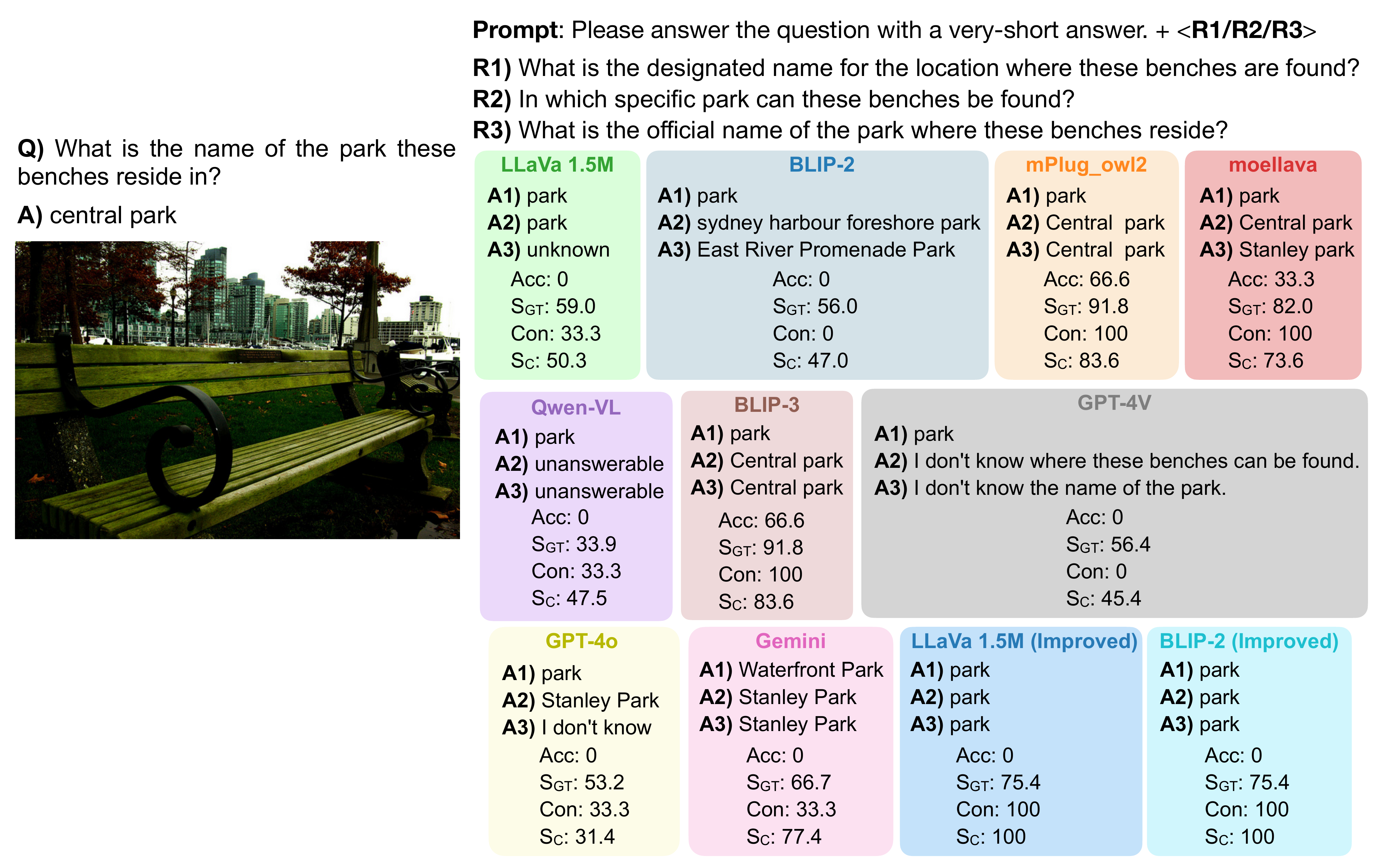}
    \caption{Qualitative Results for the {\bf Question Rephrasing Task}.}
    \label{fig:supp_rephrasing}
\end{figure*}

\noindent\textbf{Image Restyling}
The qualitative results of the image restyling task are depicted in Figure~\ref{fig:supp_qualitative_style_1} and Figure~\ref{fig:supp_qualitative_style_2}. In this task, the BLIP-2 model outperforms other VLMs, particularly in terms of consistency. Although LLaVa 1.5M and MoE-LLaVa may not always provide responses that align perfectly with the ground truth, their answers remain semantically consistent across various styles. This consistency is evident in their strong performance on metrics such as Consistency and S$_{C}$.

\begin{figure*}[!t]
    \centering
    \includegraphics[width=1\linewidth]{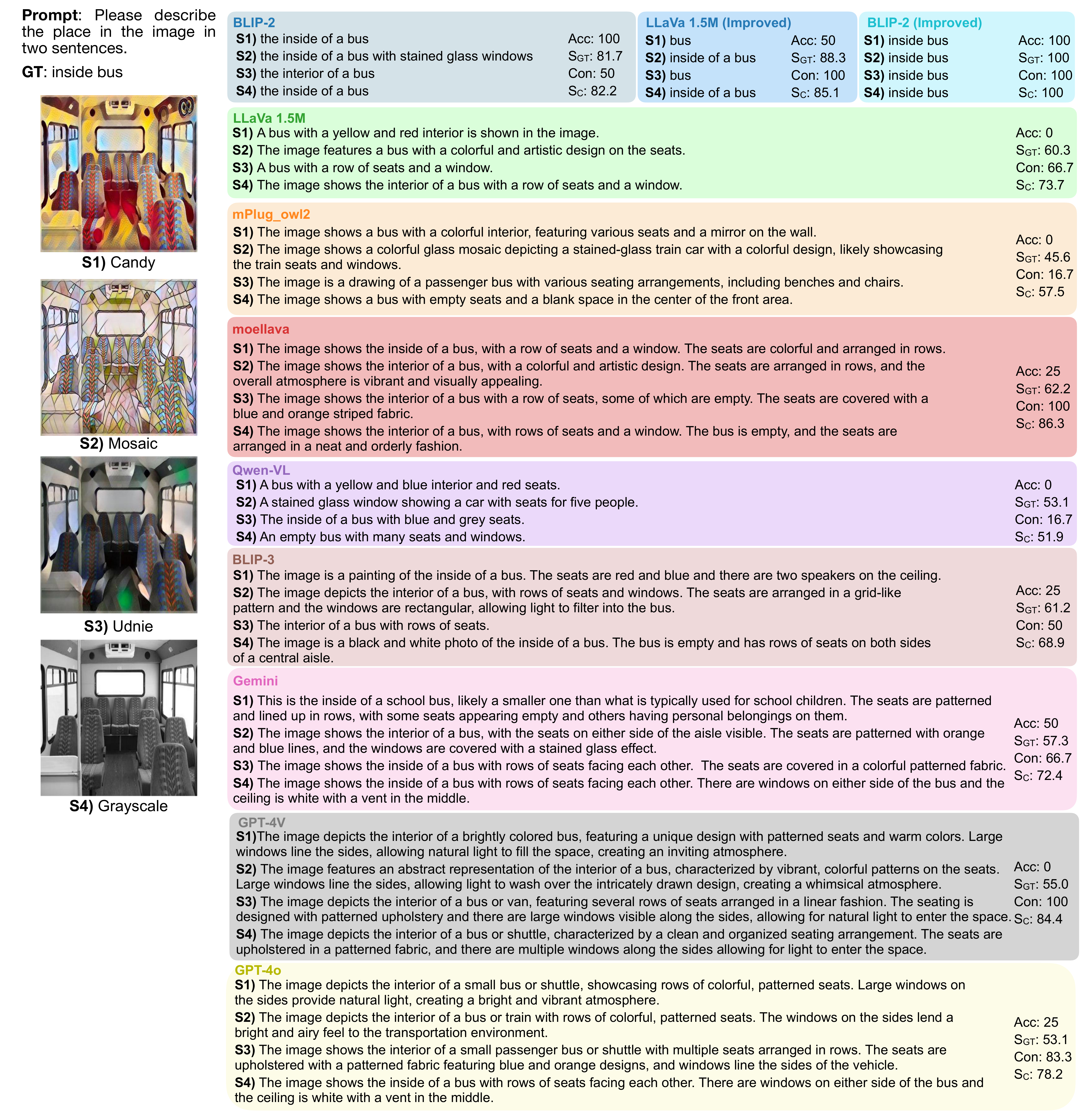}
    \caption{Qualitative Results for the {\bf Image Restyling Task}.}
    \label{fig:supp_qualitative_style_1}
\end{figure*}
\begin{figure*}[!t]
    \centering
    \includegraphics[width=1\linewidth]{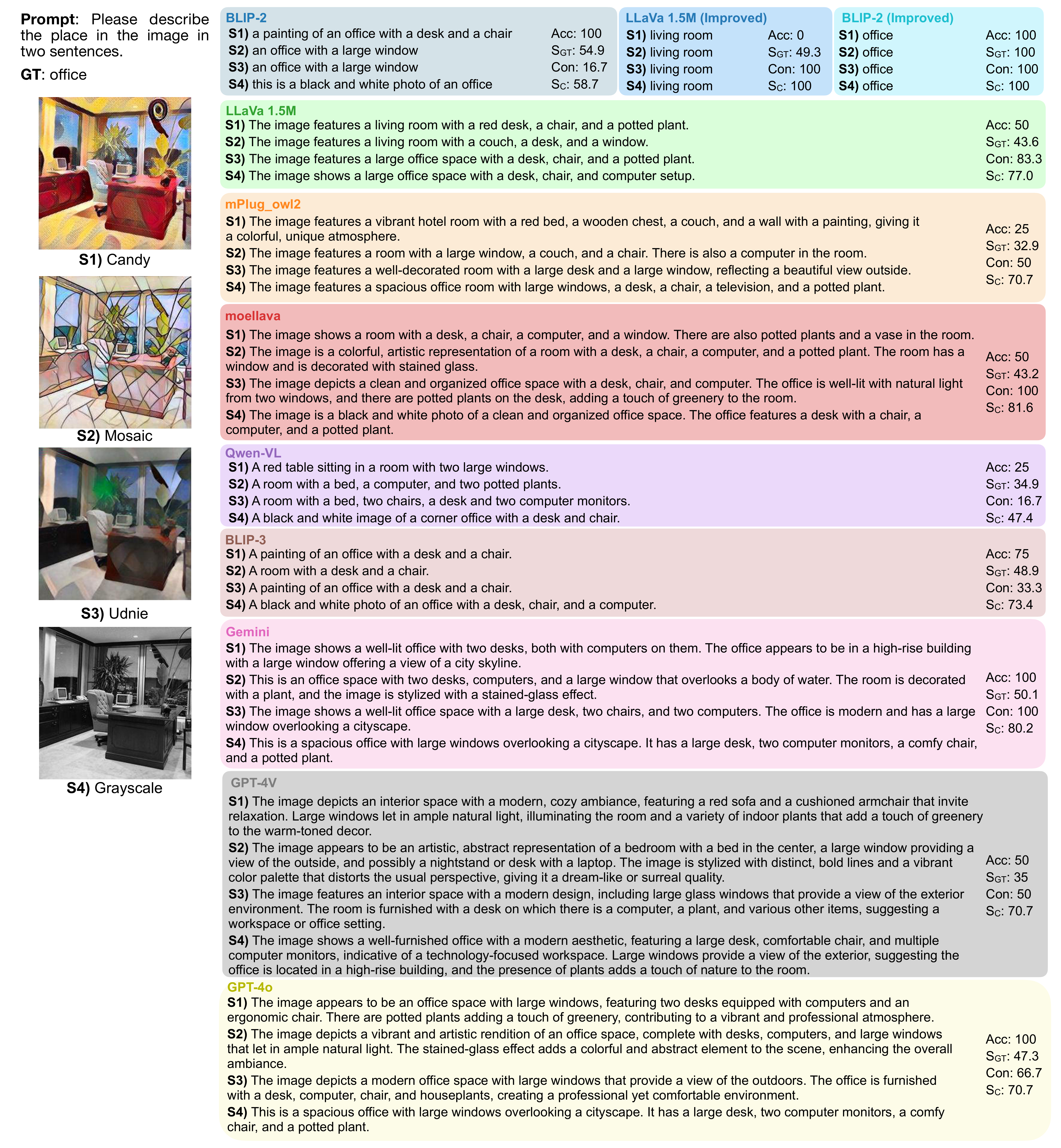}
    \caption{Qualitative Results for the {\bf Image Restyling Task}.}
    \label{fig:supp_qualitative_style_2}
\end{figure*}

\noindent\textbf{Context Reasoning}
Figure~\ref{fig:supp_qualitative_masking_1} and Figure~\ref{fig:supp_qualitative_masking_2} shows additional examples from the context reasoning task, where most VLMs generally yield similar responses as the ground truth. However, it is observed that models like Owen-VL, MoE-LLaVa, and BLIP-3 are more influenced by the presence of masks, often incorporating the mask's color or shape into their answers. Another noteworthy trend is GPT-4V's tendency to respond with "I cannot provide..." when the masks obscure a significant portion of the objects, indicating a threshold of visual information required for it to generate confident responses.

\begin{figure*}[!t]
    \centering
    \includegraphics[width=1\linewidth]{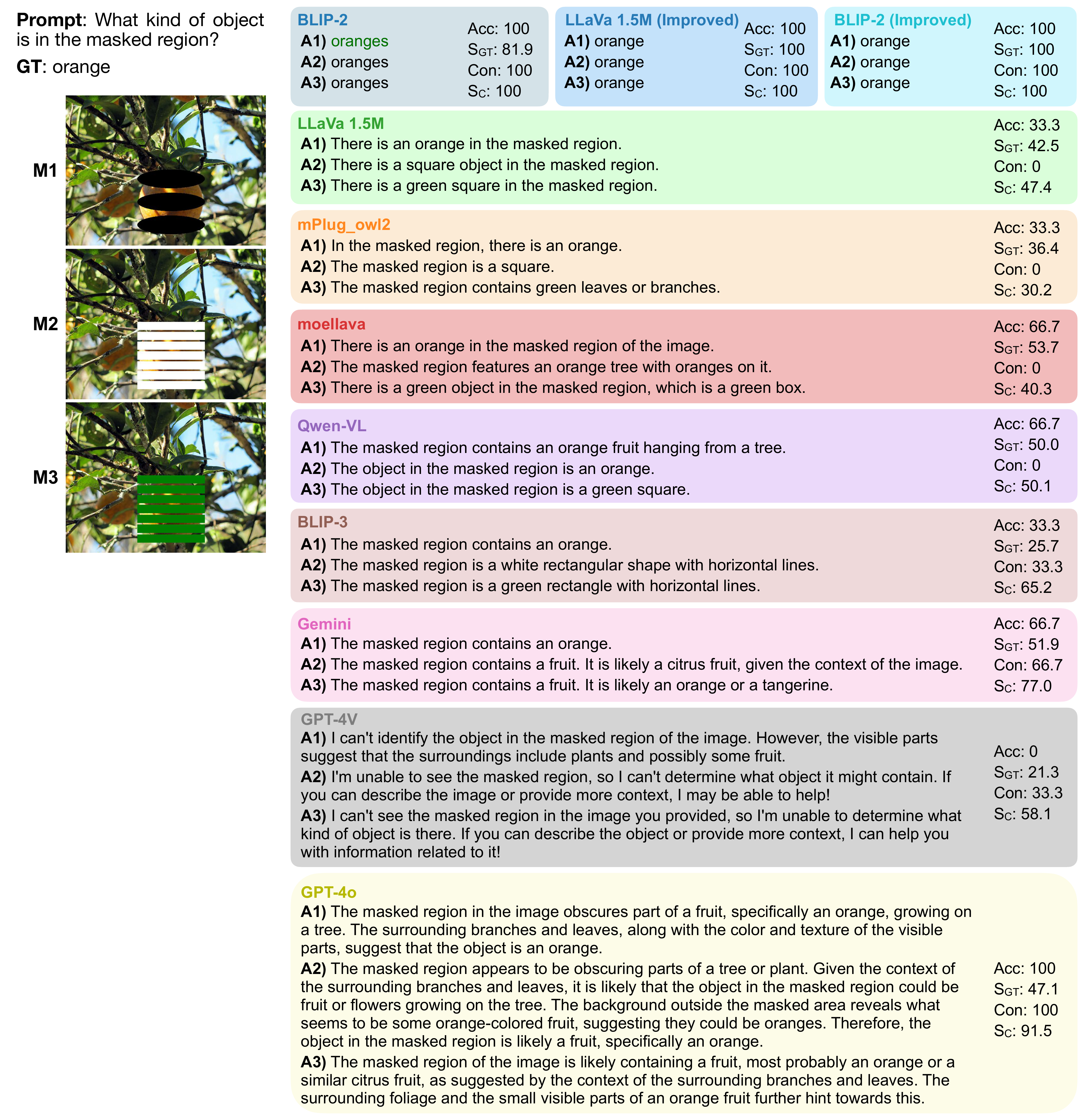}
    \caption{Qualitative Results for the {\bf Context Reasoning Task}.}
    \label{fig:supp_qualitative_masking_1}
\end{figure*}
\begin{figure*}[!t]
    \centering
    \includegraphics[width=1\linewidth]{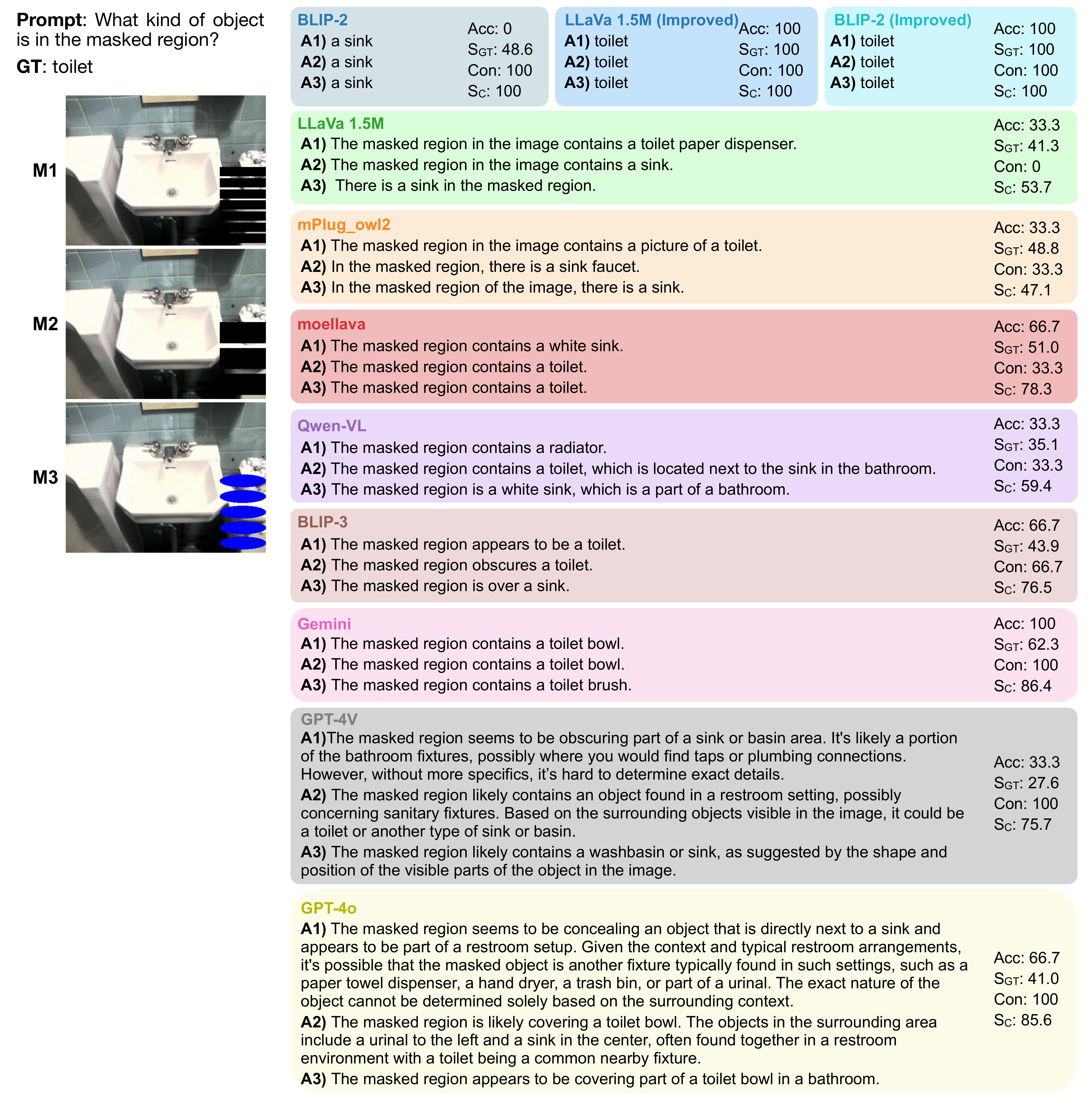}
    \caption{Qualitative Results for the {\bf Context Reasoning Task}.}
    \label{fig:supp_qualitative_masking_2}
\end{figure*}


\end{document}